\documentclass[10pt,twocolumn,letterpaper]{article}

\usepackage{iccv}
\usepackage{times}
\usepackage{epsfig}
\usepackage{graphicx}
\usepackage{amsmath}
\usepackage{amssymb}
\usepackage{multirow}
\usepackage{multicol}
\usepackage{bbm}
\usepackage{float}
\usepackage{booktabs}
\usepackage[accsupp]{axessibility} 

\newtheorem{theorem}{Theorem}

\newtheorem{definition}{Definition}
\newtheorem{assumption}{Assumption}
\newtheorem{remark}{Remark}

\usepackage[pagebackref=true,breaklinks=true,letterpaper=true,colorlinks,bookmarks=false]{hyperref}
\usepackage[capitalize]{cleveref}
\crefname{section}{Sec.}{Secs.}
\Crefname{section}{Section}{Sections}
\Crefname{table}{Table}{Tables}
\crefname{table}{Tab.}{Tabs.}
\iccvfinalcopy 

\def\iccvPaperID{5199} 

\ificcvfinal\fi

\begin{document}

\title{Diverse Cotraining Makes Strong Semi-Supervised Segmentor}

\author{Yijiang Li$^1$$^\dag$\quad Xinjiang Wang$^{2}$\quad Lihe Yang$^3$$^\dag$\quad Litong Feng$^2$\quad Wayne Zhang$^2$\quad  Ying Gao$^4$ \\\newline
{\small $^1$Johns Hopkins University\quad 
$^2$SenseTime Research\quad 
$^3$The University of Hong Kong\quad 
$^4$South China University of Technology} \\
{\small\texttt{yli556@jhu.edu, wangxinjiang@sensetime.com}} }

\maketitle
\ificcvfinal\thispagestyle{empty}\fi
\def\thefootnote{\dag}\footnotetext{Work done during internship at SenseTime.}\def\thefootnote{\arabic{footnote}}

\begin{abstract}
   Deep co-training has been introduced to semi-supervised segmentation and achieves impressive results, yet few studies have explored the working mechanism behind it. In this work, we revisit the core assumption that supports co-training: multiple compatible and conditionally independent views. By theoretically deriving the generalization upper bound, we prove the prediction similarity between two models negatively impacts the model's generalization ability. However, most current co-training models are tightly coupled together and violate this assumption. Such coupling leads to the homogenization of networks and confirmation bias which consequently limits the performance. To this end, we explore different dimensions of co-training and systematically increase the diversity from the aspects of input domains, different augmentations and model architectures to counteract homogenization. Our \textit{Diverse Co-training} outperforms the state-of-the-art (SOTA) methods by a large margin across different evaluation protocols on the Pascal and Cityscapes. For example. we achieve the best mIoU of 76.2\%, 77.7\% and 80.2\% on Pascal with only 92, 183 and 366 labeled images, surpassing the previous best results by more than 5\%. Code will be available at \url{https://github.com/williamium3000/diverse-cotraining}.
\end{abstract}
\section{Introduction}
\label{sec:intro}
Deep learning has demonstrated impressive success in various applications \cite{krizhevsky2017imagenet, resnet, frcnn, huang2023multi, yang2023diffusion, yang2022deep}. However, such SOTA performance heavily depends on supervised learning which requires expensive annotations. Particularly, labeling images for semantic segmentation is much more laborsome and time-consuming compared with that of image classification \cite{cordts2016cityscapes}. 
Therefore, how to leverage the unlabeled images available in much larger quantities to improve the segmentation performance becomes crucial. Semi-supervised segmentation is thus proposed to alleviate the expensive annotation and is attracting growing attention \cite{souly2017semi}.

One line of research in semi-supervised segmentation is co-training, which was first proposed by \cite{blum1998combining} with two compatible and independent views to guarantee the theoretical learnability \cite{nigam2000analyzing}. It was first introduced to semi-supervised segmentation for cross pseudo supervision (CPS) \cite{cps}. Since most computer vision tasks provide only one view (RGB image) for each sample, CPS adopts two networks with identical architectures and different initializations to provide different opinions. 
Later researchers improve upon CPS through one or multiple extra networks \cite{filipiak2021n}, additional consistent constraints \cite{ke2020guided}, multiple heads with a shared backbone \cite{fan2022ucc, cct} and EMA teachers \cite{xiao2022semi}. Despite these new variants of co-training, few studies have discussed the working mechanism behind the remarkable performance of co-training in semi-supervised segmentation. In this paper, we revisit the assumptions behind co-training: \textit{two or multiple independent views compatible with the target function.} By deriving the generalization upper bound of Co-training, we theoretically show that the homogenization of networks accounts for the generalization error of Co-training methods. Empirically, we examine the existing co-training architectures and discover that they provide insufficient diversity (only by different initialization). We argue that similar decision boundaries or predictions will further lead to confirmation biases \cite{arazo2020pseudo, ouali2020overview} as no additional information or correction is induced. Given this problem, a natural question emerges, \textit{how to create two or multiple views that are mutually independent?} To answer this question, we systematically explore different dimensions of the Co-training process to increase diversity. Specifically, both RGB and frequency domain are adopted as two inputs that cater to different properties of an image. Different augmentations of the same image first provide distinct views for each model. Different architectures including CNN and Transformer-based networks are then demonstrated to be quite effective in co-training due to the diverse inductive biases. 
Combining these findings, we propose our holistic approach: \textit{Diverse Co-training}. \textit{Diverse Co-training}. To summarize, we make the following contributions:
\begin{itemize}
    \vspace{-2.5mm}
    \item We theoretically prove that the homogenization of networks accounts for the generalization error of Co-training and discover the lack of diversity in current co-training methods that violate the assumptions.
    \vspace{-1.5mm}
    \item We comprehensively explore the different dimensions of co-training to promote diversity including the input domains, augmentations and architectures and demonstrate the significance of diversity in co-training.
    \vspace{-1.5mm}
    \item We propose a holistic framework combining the above three techniques to increase diversity and discuss two variants with high empirical performances.
\end{itemize}


\section{Related Work}
\label{sec:related}
\textbf{Semi-supervised Learning.}
Early works introduce self-training to tackle semi-supervised learning with an iterative EM algorithm \cite{nigam2000text, dempster1977maximum, vittaut2002learning}. Instead of labeling the unlabeled data before training, consistency regularization typically enforce invariance to perturbations on the unlabeled data in an online manner \cite{laddernetwork, temporalensembling, uda, mixmatch, vat, remixmatch, sohn2020fixmatch, wang2023consistent}. Along this line of studies, researchers notice the significance of strong data augmentation. Combining with EMA Teacher \cite{meanteacher}, the "Teacher-Student" paradigm emerges. 
However, this framework suffers from the coupling problem since the teacher comes form the aggregation of student parameters \cite{ke2019dual}, which fails to transfer meaningful knowledge and further leads to confirmation biases \cite{meanteacher, ouali2020overview}. We refer to \cite{ke2019dual} for detailed analysis. To this end, Deep Co-training is proposed \cite{qiao2018deep, ke2019dual}. Initially, co-training is proposed to solve the semi text classification problem with two models and two views
\cite{blum1998combining}. The author proves the learning ability with PAC framework \cite{valiant1993view} with the assumption that two views are compatible and independent given the class \cite{nigam2000analyzing}. Unfortunately, most tasks in computer vision provides only single view for each sample (\ie RGB image) \cite{coco, imagenet, voc2012, cordts2016cityscapes}. To simulate the condition required by co-training, \cite{democratic, ke2019dual} propose to use different initialization while Tri-training \cite{tri-training} creates diverse training sets by injecting noise to true labels. Deep Co-training proposes a novel view difference constraint and adversarial examples as an additional pseudo view \cite{qiao2018deep}. Other methods such as resampling \cite{bauer1999empirical}, bagging \cite{sutton2005classification} or bootstrapping \cite{freedman1981bootstrapping} is also used to generate pseudo views. Deep Co-training is closely related to our work. Despite the two views increases diversity, adversarial examples are generated from the the original sample leading to large dependence. Moreover, models trained with adversarial examples suffers from a degraded performance \cite{tsipras2018robustness, su2018robustness} resulting to unequal roles where the original model serves as teacher and the model trained on adversarial examples the student (since original model is better with higher performance).

\textbf{Semi-supervised Segmentation} evolves from the early GAN-based methods \cite{goodfellow2020generative, souly2017semi, hung2018adversarial, mittal2019semi} which leverage the discriminator \cite{gan} to provide an auxiliary supervision for unlabeled images to simpler training paradigms of consistency regularization \cite{french2019semi, ke2020guided, kim2020structured, hu2021semi, wang2022semi, ouali2020semi, zou2020pseudoseg} and entropy minimization \cite{ke2022three, he2021re, yuan2021simple, yang2022st++}. 
 CPS first introduces co-training to provide cross supervision by using two identical networks with different initialization \cite{cps}. n-CPS builds upon CPS and propose to add additional models with different initialization \cite{filipiak2021n} while \cite{xiao2022semi} leverages the EMA of each model to teach the other model acting as teacher. Another variant \cite{fan2022ucc, cct} of Co-training is the form of shared backbone and two heads, which is parameter and computation efficient. The co-training between CNN and transformer is also explored \cite{transformer-cnn-cohort, cnn-transformer-medical} but focuses mainly on the powerful representation ability brought by the transformer. \cite{transformer-cnn-cohort} propose to distill the feature maps between the CNN and transformer models leading to more coupled networks. Our work, on the contrary, explores the diversity presented in different architectures and demonstrate that such diversity along can achieve the SOTA performance.

\textbf{Dense Vision Transformer.}
Recent works starting with ViT \cite{dosovitskiy2020image} prove the
transformer’s adaptability in CV.
Later works such as Swin Transformer \cite{liu2021swin, liu2022swin} and PVT \cite{wang2021pyramid, wang2022pvt} proves its superiority over CNN with different inductive biases trained \cite{han2021transformer, he2022masked, yuan2021tokens, touvron2021training}. Recent work also demonstrates that transformer outperforms traditional CNN on dense predictions tasks such as object detection \cite{detr, zhu2020deformable, chen2021pix2seq, li2022exploring} and semantic segmentation \cite{strudel2021segmenter, yuan2019segmentation, xie2021segformer, zheng2021rethinking}. SETR first introduce transformer to extract feature for segmentation \cite{zheng2021rethinking}. Segmenter \cite{strudel2021segmenter} leverages mask transformer to dynamically generate class masks while SegFormer \cite{xie2021segformer} designs a novel transformer backbone with pyramid structures and simple multi-layer perceptron (MLP) head for aggregation of the multi-scale features.
\begin{figure*}[htbp] 
\centering
\includegraphics[width=0.9\textwidth]{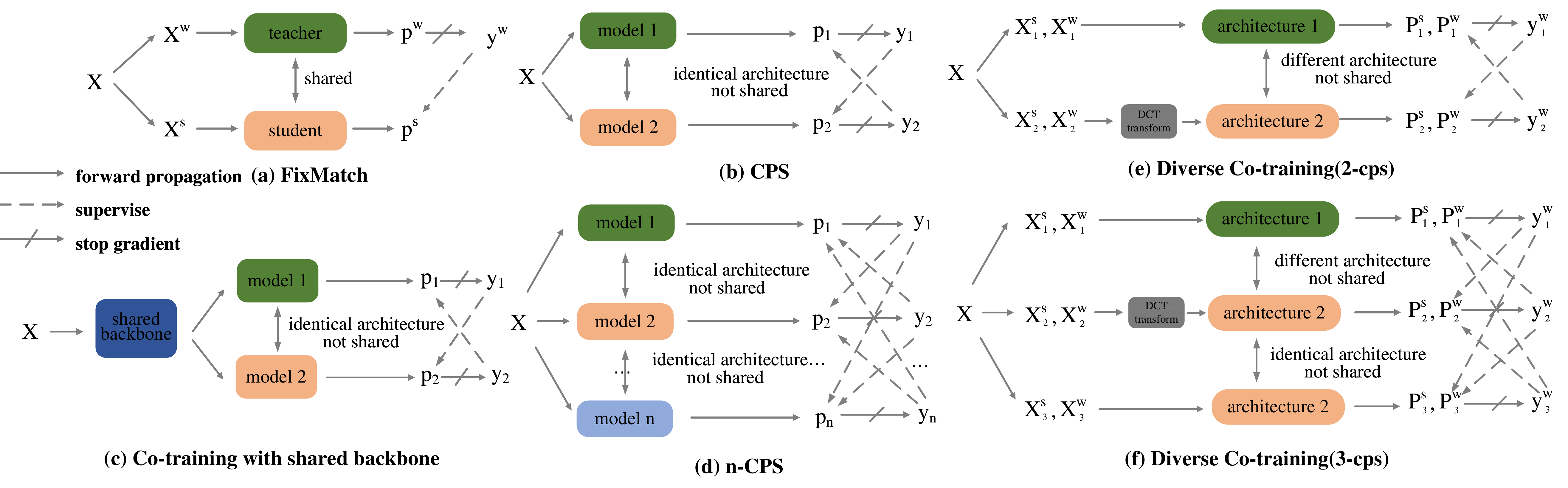}
\caption{Illustrating the architectures for (a) FixMatch \cite{sohn2020fixmatch, yang2022revisiting}, (b) CPS \cite{cps}, (c) cross heads with shared backbone \cite{fan2022ucc, cct} (d) n-CPS \cite{filipiak2021n}, (e) \textit{Diverse Co-training (2-cps)} and (f) \textit{Diverse Co-training (3-cps)}.} 
\label{fig:diff_arch} 
\end{figure*}

\section{Theoretical Analysis and Motivation}
In this section, we first provide preliminary knowledge on co-training. Then a theoretical analysis on Co-training is provided to show the relation between homogenization and generalization bound. Thirdly, we conduct a thorough investigation of the existing variants of co-training and discover that they suffer significantly from the homogenization problem, which is mainly caused by a lack of two independent views. These findings further motivate our work: \textit{how to approximate these assumptions by introducing more diversity into co-training?}
\subsection{Preliminary}
We describe Co-training in the context of segmentation. Segmentation network typically possesses an encoder $f$ and a decoder head $h$. As aforementioned, co-training in semi-supervised segmentation generally utilizes two segmentation networks $f_i(\cdot, \theta^e_1)$, $h_i(\cdot, \theta^d_1)$ parameterized by $\theta^e_i$, $\theta^d_i$, $i=1, 2$. Co-training simultaneously trains two models and the confident prediction from one model is used to supervise the other in sense of mutual teaching.
Specifically, each network generates pseudo segmentation confidence map $p_i$ after softmax operation and then the one-hot pseudo maps $y_i$ by argmax for each unlabeled data $x$:
$p_i = h_i(f_i(x, \theta^e_i), \theta^d_i)$. The generated pseudo maps are used to supervise the other network on unlabeled data.
\vspace{-0.1cm}
\begin{equation}
\small L^u = \frac{1}{|D_u|}\sum_{x \in D_u} \frac{1}{W \times H} \sum_{k=0}^{W \times H} CE(p_{1k}, y_{2k}) + CE(p_{2k}, y_{1k}) \notag
\end{equation}
where $D_u$ denotes the unlabeled data set, $W$ and $H$ denotes the size of input image and $k$ denotes the $k^{th}$ pixel. Cross entropy loss $CE$ is adopted here and for the rest of the paper.
For labeled data, each network is trained in a standard fully supervised manner:
\vspace{-0.1cm}
\begin{equation}
\small L^s = \frac{1}{|D_l|}\sum_{x \in D_l} \frac{1}{W \times H} \sum_{k=0}^{W \times H} CE(p_{1k}, y_k^{*}) + CE(p_{2k}, y_k^{*}) \notag
\end{equation}
where $D_l$ denotes the labeled data set and $y_k^*$ is the ground truth label for pixel $k$. Then, the overall objective function is a combination of the above two losses with a balance term $\lambda$: $L = L^s + \lambda L^u$

\subsection{Theoretical Analysis}
We provide a generalization bound in the PAC learning framework following \cite{blum1998combining} on the Co-training method with two models. We first give the definition of homogenization. We simplify notations from above and denote the model $f_i^j, i=1,2$ as the $i$th model of $j$th iteration and the optimal model as $f*$. 
\begin{definition}
\label{definition1}
We define homogenization as the similarity between two networks, which can be approximated by the percentage of the agreement (agree rate) of all pixels.
\begin{equation}
\small
H = Pr_{x \in D}\big[ f_1(x) = f_2(x) \big] = \frac{1}{HW} \sum_{i=1}^{HW} \mathbbm{1}(p_{1i}=p_{2i}) \notag
\end{equation}
\end{definition}
With different architectures, direct measures in parameter space are meaningless, we thus consider target space to quantify homogenization. Diversity is exactly the opposite $d(f_1, f_2) = Pr_{x \in D}\big[ f_1(x) \neq f_2(x) \big]$, which can be used to quantify the difference between any model. For instance, we can compute the generalization error with $d(f, f*)$. We simplify the Co-training pipelines for easy theoretical derivation. 
\begin{assumption}
At each step of optimization, pseudo labels of all unlabeled data are updated prior to optimization instead of online pseudo labeling.
\end{assumption}
\begin{assumption}
At each step of optimization, all unlabeled data is used except for the first step where only labeled data is used to get the initial model $f_1^0, f_2^0$.
\end{assumption}
With the PAC learning framework and the two assumptions, we can extend the generalization bound of \cite{wang2007analyzing} to iterative Co-training instead of one-step optimization and obtain the following.
\begin{theorem}
\label{theorem1}
Given hypothesis class $\mathcal{H}$ and labeled data set $D_l$ of size $l$ that are sufficient to learn an initial segmentor $f_i^0$ with an upper bound of the generalization error of $b_i^0$ with probability $\delta$ (\ie $l\ge \max\{\frac{1}{b_i^0}\ln \frac{|\mathcal{H}|}{\delta} \}$), we use empirical risk minimization to train $f_i^0$ on the combination of labeled and unlabeled set $\sigma^i$ where pseudo label are provided by the other model $f_{3-i}^0$. Then we have
$$ Pr\big[ d(f_i^k, f*) \ge b_i^k \big] \le \delta $$
if $lb_{i}^0 \le e \sqrt[M]{M!} - M$, where $M = u b_{3-i}^0$ and $b_i^k = \max \{ \frac{lb_i^0 + ub_{3-i}^0 - ud(f_{3-i}^{k-1}, f_{i}^{k}) }{l}, 0 \}$.
\end{theorem}
Theorem \ref{theorem1} shows that the bigger the difference between the two models $f_{3-i}^{k-1}$ and $f_{i}^{k}$, the smaller the upper bound of the generalization error. Thus from Theorem \ref{theorem1} and Definition \ref{definition1}, we can conclude.
\begin{remark}
Homogenization negatively impacts the generalization ability of the Co-training method leading to sub-optimal performance.
\end{remark}
With the condition that the difference between the two models is large enough $d(f_{3-i}^{k-1}, f_{i}^{k}) \ge b_{3-i}^0$, we can see that the larger the $u$ the smaller the upper bound of the generalization error. Then we have Remark \ref{remark2}.
\begin{remark}
\label{remark2}
Given a large difference between the two models, more unlabeled data decreases the generalization error of Co-training.
\end{remark}
This remark is consistent with empirical results that more unlabeled data leads to better performance. Further, with this remark, we provide theoretical guarantees for strong augmentations used in our method.
\subsection{Homogenization in Co-training}
\label{sec:limitations}
\begin{figure}[htbp] 
\centering

\includegraphics[width=0.45\textwidth]{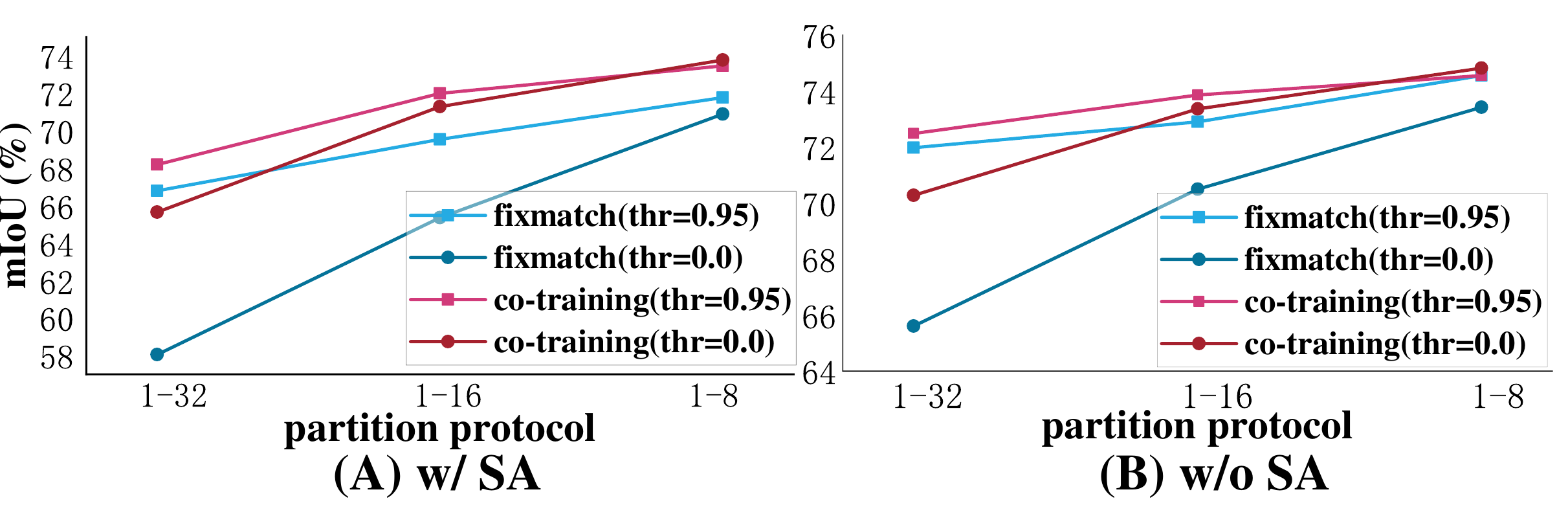}
\caption{Comparison between FixMatch and co-training under different partition protocols and thresholds (a) with or (b) without strong augmentation (SA). Best viewed in color.} 
\label{fig:comparison_fix_co} 
\end{figure}

Given the statement that homogenization negatively impacts performance, we now investigate the existing Co-training methods. Other than CPS ((b) of Figure \ref{fig:diff_arch}), we summarize two co-training paradigms. co-training with cross heads and shared backbone is widely adopted in previous works \cite{fan2022ucc, cct}, as shown in (c) of Figure \ref{fig:diff_arch}. n-CPS \cite{filipiak2021n} leverages multiple models to perform co-training, which can be seen as a generalized CPS ((d) of Figure \ref{fig:diff_arch}). We also display the paradigms used in FixMatch in (a) of Figure \ref{fig:diff_arch}. Co-training is first introduced for its merit of providing decoupled models for cross-supervision, which can alleviate confirmation biases and generate additional information for its counterpart \cite{ke2019dual}. We demonstrate the superiority of co-training over FixMatch in Figure \ref{fig:comparison_fix_co}, from which co-training outperforms FixMatch consistently on all partitions and thresholds. 

\begin{figure}[htbp] 
\centering
\includegraphics[width=0.45\textwidth]{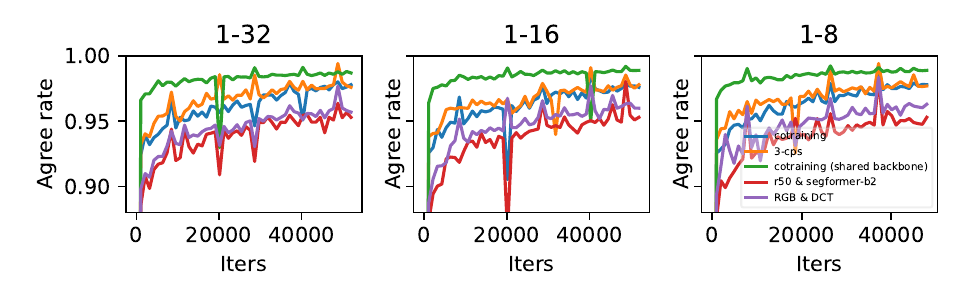}
\caption{Homogenization problem (measured by agree rate) with different partitions. 3-cps refers to n-cps when $n=3$.} 
\label{fig:agree_rate} 
\end{figure}

Despite the benefit, current deep co-training strongly violates the second assumption of co-training, as discussed in Section \ref{sec:intro}, since only a single view is used. The second assumption can weekly fulfilled through different initializations but it contributes too little difference for the two models to learn distinct decision boundaries. 
As shown in Figure \ref{fig:agree_rate}, we can observe that all three paradigms have a severer homogenization issue. We also provide rigorous analysis in logits and prediction space with L2 distance and KL Divergence demonstrating similar phenomena in Appendix \ref{app:homogenization}.

Specifically, we find that co-training with a shared backbone has the most severe homogenization as the shared backbone delivers the same features for each head. n-CPS also suffers from a severer homogenization problem than CPS because cross-supervision of a stack of models further enforces these models to predict similarly. We emphasize that agree rate alone is not sufficient to judge the effectiveness of a co-training method, as we cannot simply say a method is worse because they provide similar predictions. Thus, we empirically show in Table \ref{tab:strong_aug} that \textbf{less similar models bring performance benefits} (\ie co-training outperforms the other two consistently over all settings).
On the other hand, co-training with diverse input domains or different architectures achieves a much lower prediction similarity providing much more information to complement the counterpart. We show in Section \ref{sec:diverse_cotraining} and evaluate in Section \ref{sec:analysis_diversity} that they both bring significant empirical benefits and thus can be viewed as two relatively more independent views compatible with the target function.

\section{Diverse Co-training}
After analyzing the limitation of current co-training paradigms, we provide a comprehensive investigation of co-training to i) promote the diversity between models and ii) provide a relatively more independent pseudo view that better fits the assumption in the PAC framework. We first introduce a better co-training baseline by adopting the strong-weak augmentation. Then, we propose and analyze three techniques to better increase the diversity between models.

\begin{table}[h]
  \small 
  \centering
  \scalebox{0.9}{
  \setlength{\tabcolsep}{1.mm}{
  \begin{tabular}{cc|ccc}
    \toprule
    & \multirow{2}{*}{Method} & 1/32 & 1/16 & 1/8 \\
    & & (331) & (662) & (1323) \\
    \hline
    \multicolumn{2}{c|}{Sup Baseline} & 61.2 & 67.3 & 70.8 \\
    \hline
    \multirow{3}{*}{w/o SA} & Co-training  & \textbf{65.66} & \textbf{71.28} & \textbf{73.77} \\
    & shared backbone  & 58.97 & 65.94 & 71.25 \\
    & 3-cps  & \underline{65.41} & \underline{70.81} & $\underline{72.84}$ \\
    \hline
    
    \multirow{3}{*}{w/ SA} & Co-training  & \textbf{70.28}(\textcolor{red}{+4.62}) & \textbf{73.36}(\textcolor{red}{+2.08}) & \textbf{74.82}(\textcolor{red}{+1.05}) \\
    & shared backbone  & 69.48 & 70.16 & \underline{73.47} \\
    & 3-cps  & \underline{69.68} & \underline{71.83} & 74.36 \\
    \bottomrule
  \end{tabular}}
  }
  \small
  
  \caption{Co-training methods on ResNet50 with or without strong augmentation (SA).}
  \label{tab:strong_aug}
\end{table}
\label{sec:diverse_cotraining}
\textbf{Strong-Weak Augmentation.}
Strong-weak augmentation paradigm supervises a strongly perturbed unlabeled image $x^s$ with the pseudo label provided by its corresponding weakly perturbed version $x^w$. A better pseudo label can be obtained with $x^w$ while more efficient learning can be conducted on $x^s$ since $x^s$ expands the knowledge \cite{xie2020self, yang2022revisiting}, alleviates confirmation biases \cite{arazo2020pseudo} and enforce models with a decision boundary in the low-density regions \cite{ouali2020overview}. Theoretically, we can also see the positive effect of strong augmentations through Remark \ref{remark2} by showing that strong augmentations can potentially increase the size of unlabeled data.
We argue that the improvements brought by strong augmentation are orthogonal to that of co-training with mutual benefits. In light of this statement, we combine the strong-weak augmentation with co-training to provide a better baseline for semi-supervised segmentation. Formally, we denote the weak augmentation $O^w$ sampled from weak augmentation space $S^w$ (\ie random cropping and flipping) and strong augmentation $O^s$ from $S^w$ (detailed in Appendix \ref{app:augmentation}). For each image $x$, we obtain the strongly augmented image $x^s = O^s(O^w(x))$ and the weakly perturbed $x^w = O^w(x)$. To combine co-training with strong-weak augmentation, each model is fed with both $x^w$ and $x^s$ and predicts on $x^w$ to supervise the other model as illustrated by (e) of Figure \ref{fig:diff_arch}. This can be formulated by replacing the $L^u$ with $L^u_{st}$, where the subscript $st$ stands for "strong-weak".
\begin{equation}
\small
L^u_{st} = \frac{1}{|D_u|}\sum_{x \in D_u} \frac{1}{W \times H} \sum_{i=0}^{W \times H} CE(p_{1i}^s, y_{2i}^w) + CE(p_{2i}^s, y_{1i}^w)\notag
\end{equation}
where $p_1^s=h_1(f_1(x^s, \theta^e_1), \theta^d_1, p_1^w = h_1(f_1(x^w, \theta^e_1), \theta^d_1))$. $p_2^s$ and $p_2^w$ are similar and thus omitted.  We empirically evaluate the effectiveness of strong-weak augmentation in combination with different co-training methods in Table \ref{tab:strong_aug}. The improvement is significant and consistent on all co-training frameworks, demonstrating our argument that co-training and strong augmentation takes effect orthogonally and complements each other. We strongly suggest taking the improved co-training as the baseline for future studies.

\begin{figure}[htbp] 
\centering
\includegraphics[width=0.45\textwidth]{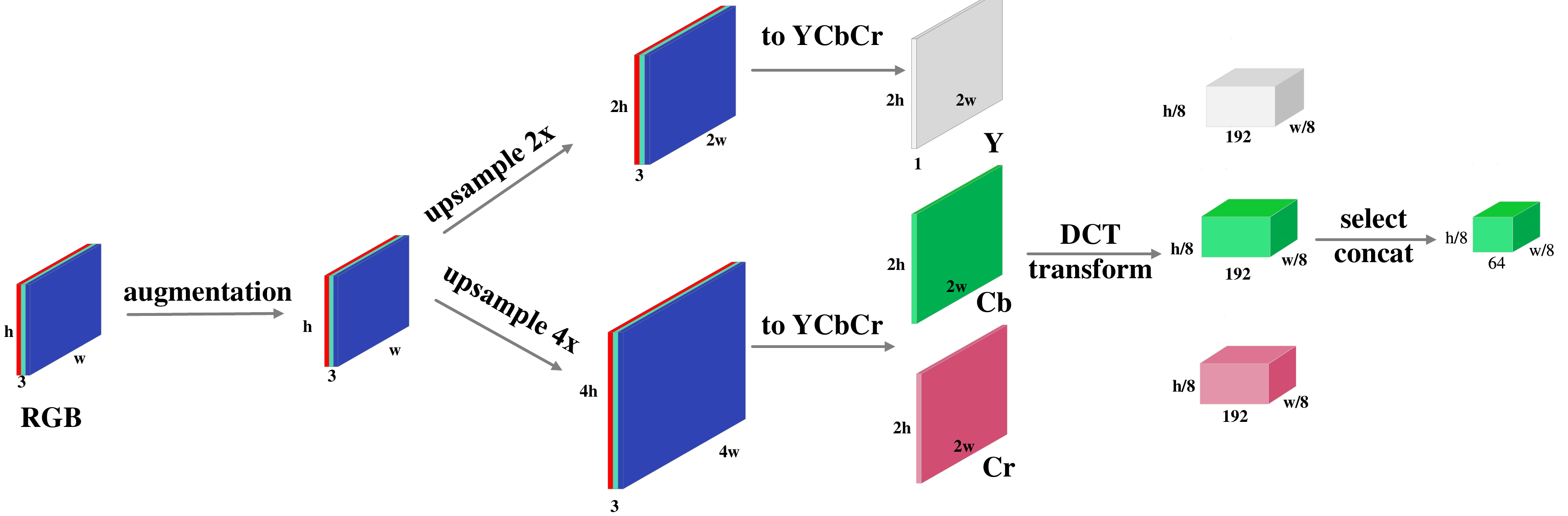}
\caption{Illustration of the DCT transform process. } 
\label{fig:dct_process} 
\end{figure}

\textbf{Diverse Input Domains as Pseudo Views.}
Co-training methods build upon two independent views while most vision tasks provide only a single view. To relax the condition, the objective is to create pseudo views with the property of i) compatibility with the target function and ii) independent with the RGB view. To this end, we propose to learn two models from different input domains. We leverage the discrete cosine transform (DCT) coefficients to generate the frequency domain input, as illustrated in Figure \ref{fig:dct_process}. We refer to Appendix \ref{app:detailed_dct} for more details. The frequency domain, in its appearance, is extraordinarily different from the RGB domain. The compression and quantization process renders the DCT relatively more independent with the RGB image than, say, an image with augmentation or adversarial perturbations. The compressed representations in the frequency domain also contain rich patterns distinct from RGB domains \cite{zhong2022detecting, xu2020learning, gueguen2018faster} which not only provides additional information to the training process but also introduce additional inductive biases in data for different perspectives in the Co-training.


\textbf{Different Augmentation Provides Different Views.}
Given a single view (\ie RGB image), augmentation is the most straightforward way to generate pseudo views \cite{he2020momentum, chen2021exploring, grill2020bootstrap, chen2020improved}. With different strong augmentation applied to the same image, distinct views can be generated so that predictions of the two models are not too similar. Particularly, random cropping has been proven effective to produce different views \cite{chen2020simple}. By randomly cropping images to different view crops, we inject view differences into the input data and Co-training is performed only on overlapping regions. Besides, random augmentations such as color jitter and greyscale also help prevent homogenization. We provide the details of augmentations used in Appendix \ref{app:augmentation}. Besides the diversity concern, different augmentations for each model also potentially increase the unlabeled data, which we show in Remark \ref{remark2} can provide a better upper bound for generalization error.

\textbf{Diverse Architecture Provides Different Inductive Biases.}
Due to the lack of two views, deep co-training utilizes different initialization to relax the condition. By taking one more step, we propose to utilize models instantiated from different architectures, \ie $f_1 \ne f_2$. In addition to different weights, diverse architectures provide different inductive biases to model the input. For the independence assumption, we provide an intuitive explanation to demonstrate why diverse architecture provides diverse views. Given the model as a function $f$ parameterized by $\theta$, it's essentially a composite function with each layer $l$ as a function $z^l = f^l(\cdot, \theta^l)$ with input from the previous layer $z^{l-1}$. By freezing the layers below some $i^{th}$ layer $f^l(\cdot, \theta^l), l=i-1, i-2, \cdots, 1$, we can view the $i^{th}$ layers and above as a trainable function that maps the output of the $i-1^{th}$ layer to the target class. Intuitively, the output of the two models from the $i-1^{th}$ layer is different and less dependent compared with the input (\ie same image) as different architectures or initialization are used. This intuitively explains why different initialization can relax the requirement of two independent views for co-training. It also explains why different architectures are better: due to the different inductive biases by different architectures, the output of every layer $i$ is much more different compared with that of different initialization, thus better fulfilling the assumption. Practically, one can leverage different CNN architectures to instantiate the two models (\eg ResNet and ResNeXT \cite{xie2017aggregated}). But to promote diversity, the co-training of CNN and transformer can provide a distinct set of inductive biases that benefit each other (\ie CNN with local modeling and transformer with the long-range dependence \cite{dosovitskiy2020image, park2022vision, raghu2021vision, wang2022anti, li2022more}).

\textbf{Holistic Approach: Diverse Co-training.}
Following the spirit of the above sections, we combine the three proposed techniques to promote a holistic framework for diverse co-training. We provide two variants of \textit{Diverse Co-training}, termed by 2-cps and 3-cps following \cite{filipiak2021n}, which leverage two models and three models to co-training respectively. Concretely, we leverage CNN and transformer as the two different architectures to maximize the discrepancy with one model trained on RGB and the other on DCT domain, as illustrated in (e) and (f) of Figure \ref{fig:diff_arch}.
The semi-supervised nature of co-training brings noise into pseudo labels for unlabeled data \cite{sohn2020fixmatch, ouali2020overview, yang2022revisiting}, thus we also provide confidence thresholding following FixMatch to filter out noisy pseudo labels in which the model has low confidence. Intuitively, each model should supervise the other model with the pseudo labels it's most confident in. We introduce $n$ ($n=2, 3$) additional scalar hyperparameters $\tau_i$ ($i=1, 2,\cdots, n$) denoting the threshold above which we retain a pseudo-label. Formally, we reformulate the unlabeled term as followed. We omit the sum over $H\times W$ for simplicity. 
\begin{gather*}
\small
L^u_{st} = \frac{1}{|D_u|}\sum_{x \in D_u} \mathbbm{1}(max(p_{2i}^w)>\tau_2) l_{ce}(p_{1i}^s, y_{2i}^w) + \\
\mathbbm{1}(max(p_{1i}^w)>\tau_1) l_{ce}(p_{2i}^s, y_{1i}^w)
\end{gather*}
Unlike other methods \cite{fan2022ucc, yuan2021simple, transformer-cnn-cohort} which either inserts modules or utilize the output of intermediate layers, we emphasize that our holistic approach leverages off-the-shelf segmentation networks without changing or probing its inside components, which can be quickly incorporated with any new SOTA segmentation methods and even be easily adapted to other fields such as semi-supervised classification and object detection.

\vspace{-3mm}
\begin{table}[H]
  \small
  \centering
  \scalebox{0.9}{
  \setlength{\tabcolsep}{1.0mm}{
  \begin{tabular}{c|c|cc}
    \toprule
    & \multirow{2}{*}{Input Domain} & 1/32 & 1/16 \\
    & & (331) & (662) \\
    \midrule
    \multirow{4}{*}{w/o SA} & RGB  & 65.66 & 71.28\\
    &DCT        & 65.33 & 67.37 \\
    &RGB \& DCT & \underline{69.45} / 69.03 & \textbf{72.46} / \underline{72.03} \\
    &RGB \& HSV & \textbf{69.65} / 67.05 & 71.74 / 69.89 \\
    \midrule
    \multirow{4}{*}{w/ SA} & RGB               & 70.28 & 73.36  \\
    &DCT               & 70.65 & 73.26 \\
    &RGB \& DCT        & \underline{71.88} / \textbf{72.00} & \textbf{74.10} / \underline{73.94}  \\
    &RGB \& HSV        & 70.40 / 68.30 & 72.64 / 70.91   \\
    \bottomrule
  \end{tabular}}
  }
  \caption{Performance of co-training with different domains. For cells with two numbers, the left one is the result of RGB model.}
  \label{tab:diff_input}
\end{table}
\vspace{-6mm}
\section{Experiment}
We conduct experiments in this section. The objective is to i) demonstrate our argument that diversity plays a crucial role in co-training ii) by illustrating that the three proposed techniques can effectively improve the performance and prevent the model from being tightly coupled with each other and iii) demonstrate the effectiveness by comparing with other state-of-the-arts.

\subsection{Experiment Setup}
\label{sec:exp_setup}
\textbf{Datasets.} We leverage two datasets for evaluating the effectiveness of our idea. \textit{PASCAL VOC 2012} \cite{hariharan2011semantic} is constructed by a combination of the Pascal dataset \cite{everingham2015pascal} with high-quality train and validation images and the coarsely annotated SBD dataset \cite{hariharan2011semantic}, resulting in a total of 10582 training images. Following prior arts, we randomly sample labeled images from i) the original high-quality training images, and ii) the mixed 10582 images. \textit{Cityscapes} \cite{cordts2016cityscapes} is an urban scene dataset with 19 classes and a total of 2975 high-resolution (2048 × 1024) training images as well as 500 validation images. We follow prior arts and divide the dataset by randomly sub-sampling 1/4, 1/8 and 1/30 of the total training images as labeled set and the rest as the unlabeled set. We crop each image to 769x769 during training.

\textbf{Evaluation.} We evaluate the segmentation performance with the mean Intersection-over-Union (mIoU) metric. For all partition protocols, we report the results on the \textit{PASCAL VOC val} set with single scale testing on origin resolution and \textit{Cityscapes val} set with single scale sliding window evaluation with crop size of 769 following \cite{yang2022revisiting, wang2022semi}

\textbf{Implementation Details.} For fair comparisons, we leverage the widely adopted DeepLabv3+ with ResNet as CNN and the SegFormer as the transformer architecture.
The backbones of both architectures are pre-trained on ImageNet 1K. We utilize the pre-trained weights on ImageNet 1K for frequency domain from \cite{xu2020learning}. During training, we leverage a batch size of 16 for Pascal and 8 for Cityscapes with a labeled-unlabeled ratio of 1. We train Pascal and Cityscapes for 80 and 240 epochs with an initial learning rate of 0.001 and 0.005 respectively and polynomial learning rate decay following \cite{cps}. 

\begin{table}[htbp]
  \small 
  \centering
  \scalebox{1.0}{
  \setlength{\tabcolsep}{1mm}{
  \begin{tabular}{c|c|cc}
    \toprule
    & \multirow{2}{*}{Backbone} & 1/32 & 1/16       \\
    &                         & (331) & (662)  \\
    \hline
    \multirow{3}{*}{w/o SA} & R50           & 65.66 & 71.28  \\
    & mit-b2        & 71.01  & 74.53 \\
    & R50 \& mit-b2 & \textbf{71.58} / \underline{71.03} & \textbf{74.94} / \underline{74.84} \\
    \hline
    \multirow{7}{*}{w/ SA} & R50 & 70.28 & 73.36  \\
    & mit-b2                     & 74.51  & \underline{75.29}  \\
    & R50 \& mit-b2 & \underline{74.85} / \textbf{74.87} & 75.12 / $\textbf{75.85}$  \\
    \cline{2-4}
    & ResNeSt50 & 70.92 & \underline{75.58} \\
    & ResNeXt50 & 71.18 & 72.77 \\
    & R50 \& ResNeST50 & \underline{72.70} / \textbf{73.56} & 73.41 / \textbf{75.65} \\
    & R50 \& ResNeXT50 & 72.15 / 72.39 & 74.41 / 74.56         \\

    \bottomrule
  \end{tabular}}
  }
  \caption{Performance of co-training with different architectures. We refer ResNet50 as R50 and SegFormer-b2 as mit-b2 \cite{xie2021segformer}. For cells with two numbers, the left one is the result of ResNet50.}
  \label{tab:diff_arch}
\end{table}

\subsection{Analysis on How to Promote Diversity}
\label{sec:analysis_diversity}
We empirically analyze the three techniques proposed with ResNet50. All experiments and figures in Section \ref{sec:limitations} and this sections are conducted on \textit{PASCAL VOC} with partition protocol 1/32 and 1/16.
\begin{figure}[htbp] 
\centering
\includegraphics[width=0.5\textwidth]{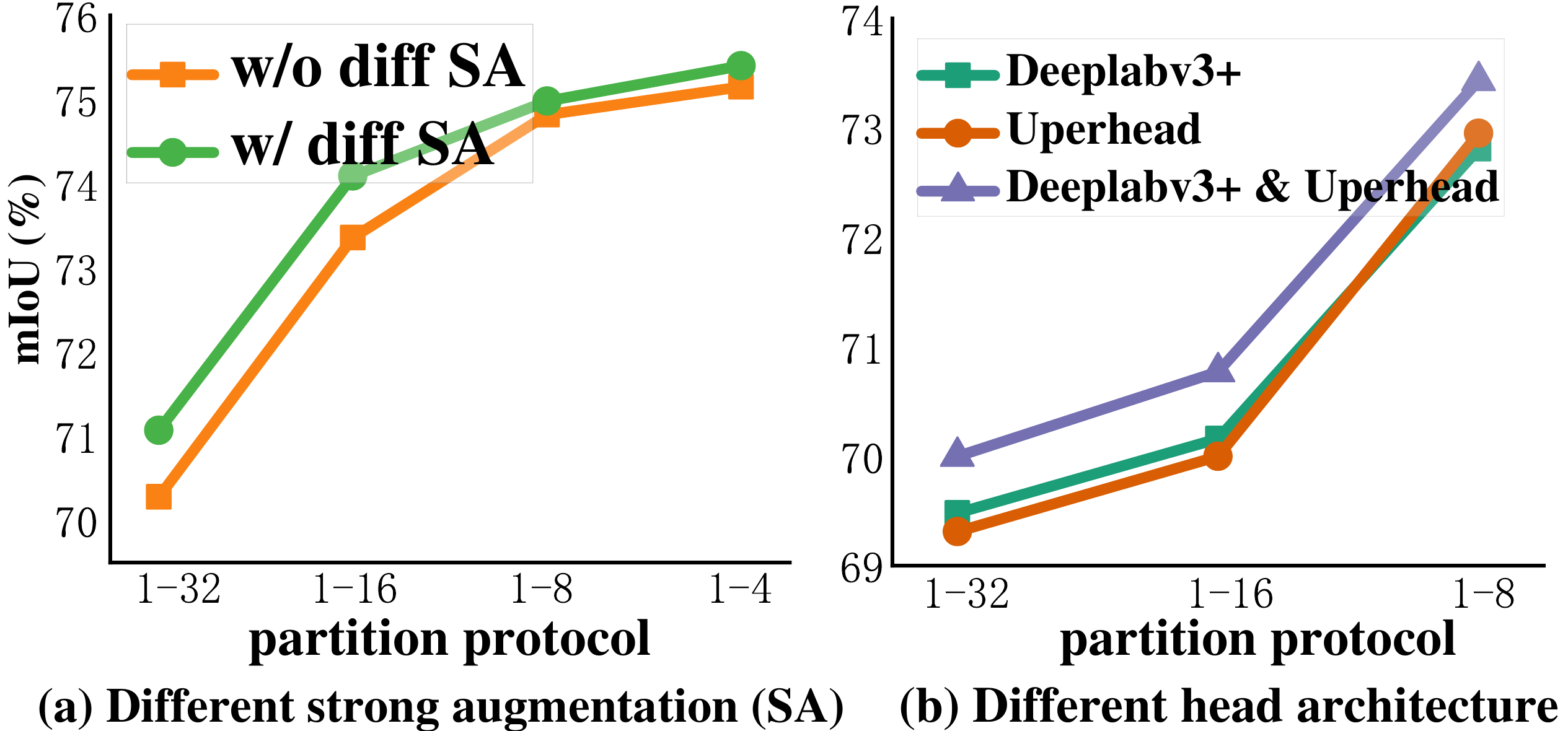}
\caption{(a) Empirical evaluation of co-training with or without different strong augmentation (SA); (b) Empirical evaluation of co-training with different head structures.} 
\label{fig:diff_SA_head} 
\end{figure}

\textbf{Different Input Domains} provides two views that are relatively independent of each other. We demonstrate that the DCT domain better fulfills the two assumptions of co-training. Firstly, we show in Table \ref{tab:diff_input} that training only on the DCT domain achieves similar accuracy as the RGB domain, demonstrating that the DCT view is compatible with the target function of the RGB domain and sufficient to train a quality segmentor. As per Figure \ref{fig:agree_rate}, co-training with different domains significantly lowers the similarity between models, illustrating that different domains increase the diversity of the two models. We further evaluate empirically on \textit{VOC PASCAL} where the RGB \& DCT outperforms the baselines with single view consistently on all settings. Notice that both models of RGB \& DCT obtain a significant improvement, demonstrating that diversity benefits two models mutually instead of a unidirectional teacher-student one. We also conduct experiments on HSV, a color space different from RGB and also observe similar but fewer improvements over baseline compared with DCT. We hypothesize this is due to the discrepancy between DCT and RGB being larger than that of HSV and RGB.
\begin{table}[h]
  \small 
  \centering
  \scalebox{1.0}{
  \setlength{\tabcolsep}{0.8mm}{
  \begin{tabular}{c|c|ccccc}
    \toprule
    Method & Resolution & 92 & 183 & 366 & 732 & 1464 \\
    \hline
    \hline
    \multicolumn{7}{c}{ResNet50}\\
    \hline
    Sup Baseline & 513x513 & 39.1  & 51.3 & 60.3 & 65.9 & 71.0\\
    PseudoSeg \cite{zou2020pseudoseg} & 512x512 & 54.9 & 61.9 & 64.9 & 70.4 & - \\
    PC$^2$Seg \cite{zhong2021pixel} & 512x512 & 56.9 & 64.6 & 67.6 & 70.9 & - \\
    \hline
    Ours (2-cps) & 513x513  & \underline{71.8} & \underline{74.5} & \textbf{77.6} & \underline{78.6} & \underline{79.8} \\
    Ours (3-cps) & 513x513  & \textbf{73.1} & \textbf{74.7} & \underline{77.1} & \textbf{78.8} & \textbf{80.2}  \\
    \hline
    \hline
    \multicolumn{7}{c}{ResNet101} \\
    \hline
    Sup Baseline & 321x321 & 44.4 & 54.0 & 63.4 & 67.2 & 71.8 \\
    ReCo \cite{liu2021bootstrapping}     & 321x321  & 64.8 & 72.0 & 73.1 & 74.7 & - \\
    ST++ \cite{yang2022st++}    & 321x321  & 65.2 & 71.0 & 74.6 & 77.3 & 79.1  \\
    \hline
    ours (2-cps) & 321x321  & \underline{74.8} & \textbf{77.6} & \underline{79.5} & \underline{80.3} & \textbf{81.7}  \\
    ours (3-cps) & 321x321  & \textbf{75.4} & \underline{76.8} & \textbf{79.6} & \textbf{80.4} & \underline{81.6}  \\
    \hline
    Sup Baseline & 512x512 & 42.3 & 56.6 & 64.2 & 68.1 & 72.0 \\
    MT \cite{meanteacher}       & 512x512 & 48.7 & 55.8 & 63.0 & 69.16 & - \\
    CPS\cite{cps}      & 512x512  & 64.1 & 67.4 & 71.7 & 75.9 & - \\
    U$^2$PL \cite{wang2022semi}   & 512x512  & 68.0 & 69.2 & 73.7 & 76.2 & 79.5 \\
    PS-MT \cite{psmt}    & 512x512  & 65.8 & 69.6 & 76.6 & 78.4 & 80.0 \\
    \hline
    ours (2-cps) & 513x513  & \textbf{76.2} & \underline{76.6} & \textbf{80.2} & \underline{80.8} & \underline{81.9} \\
    ours (3-cps) & 513x513  & \underline{75.7} & \textbf{77.7} & \underline{80.1} & \textbf{80.9} & \textbf{82.0} \\
    \bottomrule
  \end{tabular}}
  }
  \caption{Comparison with state-of-the-art methods on the \textit{Pascal} dataset. Labeled images are from the high-quality training set.}
  \label{tab:voc_183}
\end{table}
\textbf{Different Augmentation} further promotes diversity in co-training. We empirically show in (a) of Figure \ref{fig:diff_SA_head} that different augmentations for each model are effective and show consistent improvement over the baseline.


\begin{figure}[htbp] 
\centering
\includegraphics[width=0.5\textwidth]{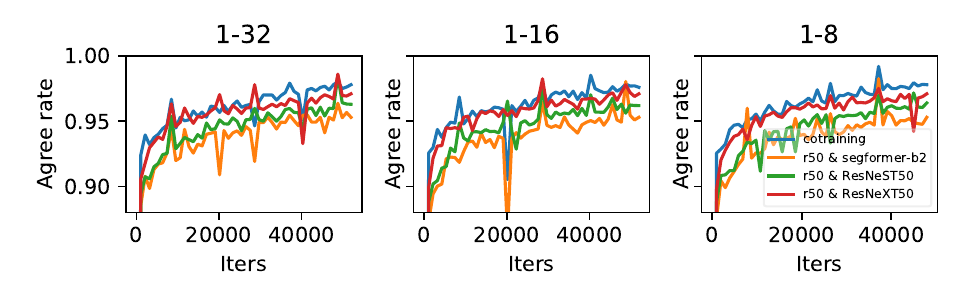}
\caption{Homogenization (measured by agree rate) with different architectures.} 
\label{fig:agree_rate_cnn} 
\end{figure}

\textbf{Different Architectures} provides different inductive biases and thus better independent views as discussed in Section \ref{sec:diverse_cotraining}. From Figure \ref{fig:agree_rate}, we already demonstrate that less similar decision boundaries can be obtained with different architectures. We here empirically evaluate the performance of co-training with different architectures, as presented in Table \ref{tab:diff_arch}. Both R50 and mit-b2 can be further improved through co-training compared with their corresponding individual baseline. For instance, we observe a remarkable improvement of ResNet50 over single R50 co-training (\eg $5.37\%$ and $3.56\%$ on 1/32 and 1/16 w/o SA), which can be attributed to the high performance of mit-b2. However, ResNet50 of R50 \& mit-b2 even surpasses the individual mit-b2 demonstrating that the cross-supervision between different architectures provides additional information other than the pseudo labels from mit-b2.
We also provide co-training with different CNN architectures and illustrate in Figure \ref{fig:agree_rate_cnn} that different CNNs are more coupled compared with CNN and transformer. Empirically, despite baseline ResNeSt50 obtaining better performance than baseline mit-b2, R50 \& mit-b2 outperforms R50 \& ResNeSt50 in all settings. We also notice the improvement of ResNet50 in co-training with different CNN architectures is less than that of R50 \& mit-b2, which further illustrates our point. To further prove the concept, we additionally conduct experiments on co-training with shared backbone and leverage different decoder head structures, as shown in (b) of Figure \ref{fig:diff_SA_head}. Co-training with DeepLabv3+ head \cite{deeplabv3plus2018} and UPerHead \cite{xiong2019upsnet} is better than any baselines with single-head architecture consistently.
\begin{table}[h]
  \small 
  \centering
  \scalebox{1.0}{
  \setlength{\tabcolsep}{0.8mm}{
  \begin{tabular}{c|c|cccc}
    \toprule
    \multirow{2}{*}{Method} & \multirow{2}{*}{Resolution} & 1/32 & 1/16 & 1/8 & 1/4 \\
    & & (331) & (662) & (1323) & (2646) \\
    \hline
    \hline
    Sup Baseline & 321x321 & 55.8 & 60.3 & 66.8 & 71.3 \\
    CAC\cite{lai2021semi}      &  320x320 & - & 70.1 & 72.4 & 74.0 \\
    ST++\cite{yang2022st++}     & 321x321  & - & 72.6 & 74.4 & 75.4  \\
    \hline
    Ours (2-cps) & 321x321  & \textbf{75.2} & \underline{76.0} & \underline{76.2} & \underline{76.5}  \\
    Ours (3-cps) & 321x321  & \underline{74.9} & \textbf{76.4} & \textbf{76.3} & \textbf{76.6} \\

    \hline
    \hline
    Sup Baseline &  513x513  & 54.1 & 60.7 & 67.7 & 71.9 \\
    CPS\cite{cps}     & 512x512  & - & 72.0 & 73.7 & 74.9  \\
    3-CPS \cite{filipiak2021n}  & 512x512  & - & 72.0 & 74.2 & 75.9  \\
    ELN \cite{kwon2022semi}    & 512x512  & - & - & 73.2 & 74.6  \\
    PS-MT \cite{psmt}   & 512x512  & - & 72.8 & 75.7 & 76.4  \\
    U$^2$PL*  \cite{wang2022semi}  & 513x513  & - & 72.0 & 75.1 & 76.2  \\
    \hline
    Ours (2-cps) & 513x513 & \textbf{75.2} & \underline{76.2} & \underline{77.0} & \underline{77.5}  \\
    Ours (3-cps) & 513x513 & \underline{74.7} & \textbf{76.3} & \textbf{77.2} & \textbf{77.7}  \\
    \bottomrule
  \end{tabular}
  }
  }
  \caption{Comparison with state-of-the-art methods with ResNet50 on the \textit{Pascal VOC 2012} dataset. Labeled images are sampled from the blended training set. The result of $U^2PL$ is reproduced with the same setting as ours.}
  \label{tab:voc_sota_r50}
\end{table}


\subsection{Comparison with State-of-the-arts}
\vspace{-2mm}
\label{sec:comp_sota}
\textbf{Pascal VOC 2012.} We only compare the most recent SOTA models due to limited space and a full comparison can be found in Appendix \ref{app:full_comparison}. We first compare the performance of \textit{Diverse Co-training} with SOTA on \textit{PASCAL VOC} on two groups of data partition protocols described in \ref{sec:exp_setup}. To ensure a fair comparison, we conduct training with resolutions of 321 and 513, which is reported with the results. On the first partition protocol, our \textit{Diverse Co-training} outperforms the prior methods by a remarkable margin, as displayed in Table \ref{tab:voc_183}. For instance, we obtain an improvement of over $10\%$ on ResNet50 compared with the PC$^2$Seg on 92, 183 and 366 partitions. With ResNet101, our method surpasses the up-to-date SOTA (\ie PS-MT) by a margin as large as $9\%$ under scarce label conditions such as 92 and 183, and outperforms all prior arts consistently on other settings. On the second protocol, as indicated by Table \ref{tab:voc_sota_r50}, our method also gains remarkable improvements over most up-to-date studies. We emphasize that our performance on 1/32 (\ie $75.2\%$) already outperforms other SOTA with 1/16 (around $72\%$) by $3\%$, which shows the effectiveness of our approach. We further report the comparison on ResNet101 and SegFormer-b3 in Appendix \ref{app:voc_r101}. \textbf{Cityscapes.} We compare the SOTA with our method in Table \ref{tab:cityscapes}. We report results on both ResNet50 \& SegFormer-b2 and ResNet101 \& SegFormer-b3. We can see that our method outperforms the current SOTA (\ie U$^2$PL) by more than $3\%$ on 1/30 and 1/8 and $1.6\%$ on 1/4 with ResNet50 \& SegFormer-b2. Similar trends can also be observed on ResNet101, where we contribute an improvement of $0.5\%$ and $1.1\%$ over PS-MT on 1/8 and 1/4 protocol.

\begin{table}[h]
  \small 
  \centering
  \scalebox{1.0}{
  \setlength{\tabcolsep}{0.5mm}{
  \begin{tabular}{c|ccc|c|ccc}
    \toprule
    \multirow{3}{*}{Method} & \multicolumn{3}{c|}{ResNet50} & \multirow{3}{*}{Method} & \multicolumn{3}{c}{ResNet101} \\
    \cline{2-4}
    \cline{6-8}
    & 1/30 & 1/8 & 1/4 &       & 1/16  & 1/8 & 1/4 \\
    & (100) & (372) & (744)&     & (186) & (372) & (744) \\
    \hline
    \hline
    Sup Baseline   & 54.8  & 70.2 & 73.6 & Sup Baseline & 66.8 & 72.5 & 76.4\\
    CAC \cite{lai2021semi} & -  & 69.7 & 72.7 & CutMix \cite{french2019semi} & 67.9 & 73.5 & 75.4 \\
    CPS \cite{cps} & -  & 74.4 & 76.9  & CPS \cite{cps}& 70.5 & 75.7 & 77.4 \\
    ST++ \cite{yang2022st++}     & 61.4 & 72.7 & 73.8 & U$^2$PL \cite{wang2022semi} & 74.9 & 76.5 & 78.5 \\
    U$^2$PL* \cite{wang2022semi} & 59.8 & 73.0 & 76.3 & PS-MT \cite{psmt} & - &76.9 & 77.6 \\
    \hline
    Ours (2-cps)  & \underline{64.5}  & \underline{76.3} & \underline{77.1} & Ours (2-cps) & \underline{75.0}  & \underline{77.3} & \textbf{78.7}  \\
    Ours (3-cps)  & \textbf{65.5}  & \textbf{76.5} & \textbf{77.9} & Ours (3-cps) & \textbf{75.7}  & \textbf{77.4} & \underline{78.5} \\
    \bottomrule
  \end{tabular}
  }
  }
  \caption{Comparison with state-of-the-art methods on the Cityscapes dataset. Result of U$^2$PL on ResNet50 (marked with *) is from \cite{yang2022revisiting} which is reproduced with the same setting as ours.}
  \label{tab:cityscapes}
\end{table}

\subsection{Ablation Study and Analysis}
\label{sec:ablation}
\begin{figure}[htbp] 
\centering
\includegraphics[width=0.45\textwidth]{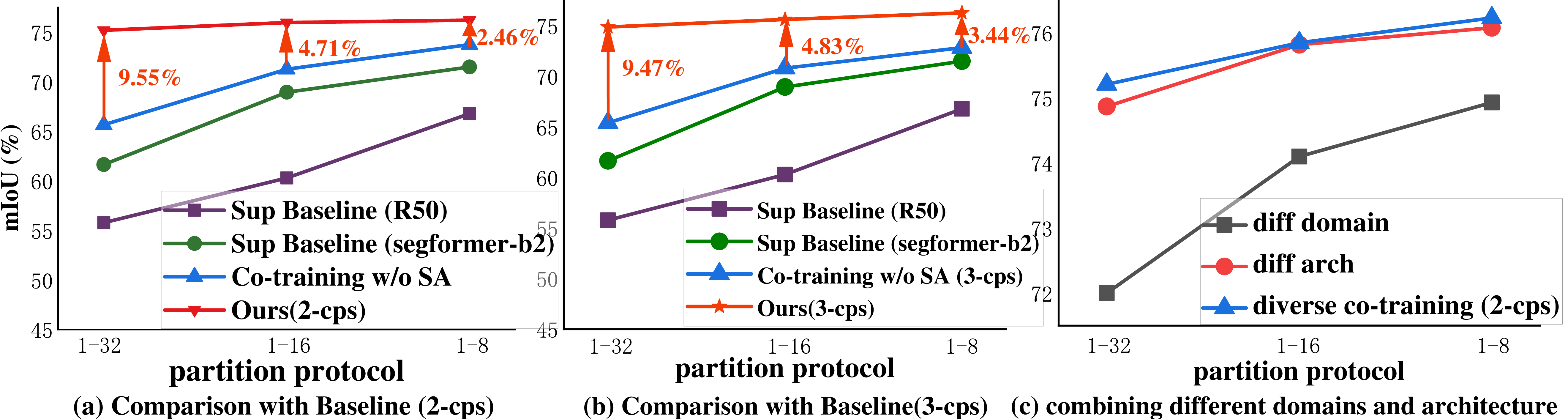}
\caption{Comparison with baseline model with (a) \textit{Diverse Co-training (2-cps)} (b) \textit{Diverse Co-training (3-cps)}; (c) Ablation study of combining diverse domains and different architectures.} 
\label{fig:comp_baseline} 
\end{figure}

\begin{figure}[h] 
\centering
\setlength{\abovecaptionskip}{-0.5mm}
\setlength{\belowcaptionskip}{-2mm}
\includegraphics[width=0.45\textwidth]{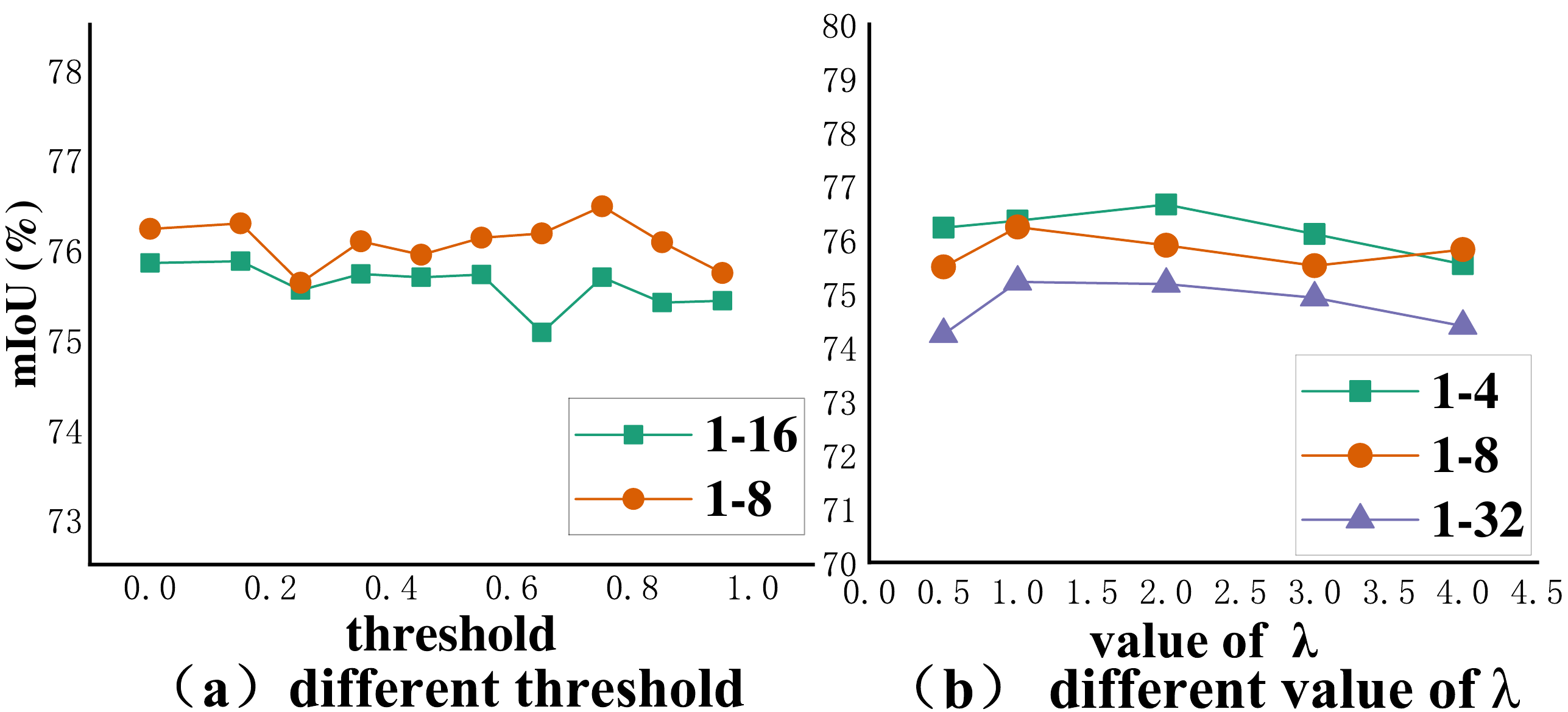}
\caption{Performance on (a) different confidence thresholds $\tau$ and (b) different weight $\lambda$.} 
\label{fig:diff_thr_weight} 
\end{figure}

\textbf{Comparison with Baseline.} We compare our methods with baseline in Figure with ResNet50 and SegFormer-b2 on \textit{PASCAL VOC}. We compare both variants of our method, \ie 2-cps and 3-cps, to each corresponding baseline (\ie CPS and n-CPS) and the labeled-only baseline. As per Figure \ref{fig:comp_baseline}, we achieve an improvement $9.55\%$, $4.71\%$, $2.46\%$ and $9.47\%$, $4.83\%$, $3.44\%$ over their corresponding baseline respectively. The improvement over the supervised baseline is larger. It's worth mentioning that the n-CPS baseline with single architecture suffers from homogenization which consequently limits the performance compared with the CPS baseline (shown in Table \ref{tab:strong_aug}). With the diversity boosted, \textit{Diverse Co-training} (3-cps) now outperforms both baselines by a large margin, further demonstrating that diversity is crucial in co-training.
Co-training is similar to knowledge distillation (KD) in the sense that they both possess a teaching process, the difference lies in that the teacher in KD is usually fixed and teaching is unidirectional. We provide a KD baseline comparison in Appendix \ref{app:kd}.
\textbf{Combining Diverse domains and Different Architectures} renders model less coupled further increasing the diversity and encouraging the exploration at early stage of training. Empirically, we also demonstrate a improvement of the combination over the each individual, as shown in (c) of Figure \ref{fig:comp_baseline}. We also provide a more complete ablation study on each one of the component and their combinations in Appendix \ref{app:more_ablation}.
\textbf{Confidence Threshold $\tau$ and Unlabeled Weight $\lambda$.} As proposed in Section \ref{sec:diverse_cotraining}, our holistic \textit{Diverse Co-training} leverages confidence threshold $\tau$ to retrain confident samples as pseudo label and balance the losses on unlabeled data with weight $\lambda$. To be as simple as possible, we set the threshold $\tau =0.0$ and $\lambda =1.0$ as default for the experiments reported in this paper. But we also report the performance of different values in Figure \ref{fig:diff_thr_weight}. We emphasize that a better performance can be obtained if optimal hyperparameters is thoroughly searched on each setting as in \cite{psmt}, which is omitted due to computation reason.
\textbf{Number of parameters.} We report the parameters of all backbones 
used in Section \ref{sec:analysis_diversity} in Appendix \ref{app:param}. We also provide a rigorous analysis to show that our improvement is not trivial by adding more parameters.
\vspace{-4mm}
\vspace{-4mm}
\vspace{-4mm}
\section{Conclusion}
\vspace{-2mm}
In this paper, we revisit the two core assumptions behind the deep co-training methods in semi-supervised segmentation and provide a theoretical upper bound over the generalization error that links with the homogenization of the two networks. We discover that the existing co-training paradigms suffer from severe homogenization problems and by exploring different
dimensions of co-training and systematically increasing the
diversity from three aspects, we propose a holistic framework: \textit{Diverse Co-training} which achieves remarkable improvement over previous best results on all partitions of the \textit{Pascal} and \textit{Cityscapes} benchmarks.

{\small
\bibliographystyle{ieee_fullname}
\bibliography{egbib}
}

\appendix
\clearpage
\newpage
\section{Proof of Theorem \ref{theorem1}.}
First, we can show with PAC learning \cite{mohri2018foundations} that with labeled data set $D_l$ of size $l$ where $l\ge \max\{\frac{1}{b_i^0}\ln \frac{|\mathcal{H}|}{\delta} \}$,
the generalization error of the initial segmentor $f_i^0$ is bounded by $b_i^0$ with probability $\delta$, which is a standard PAC supversied learning problem. Then, without loss of generality, we show the probability that the generalization error of $f_2^k$ denoted by $d(f_2^k, f*)$ is larger than $b_i^k$ is at most $\delta$.

we analyze the prediction difference between the segmentor $f_2^k$ and the total dataset which, at the $k$th iteration, contains the labeled set and the unlabeled set annotated by the previous segmentor $f_1^{k-1}$. We denote this dataset as $\sigma_2$.
$$d(f*, \sigma_2) = \frac{u \times d(f_1^{k-1}, f*)}{l + u}$$
$$d(f_2^k, \sigma_2) = \frac{l\times d(f_2^k, f*) + u\times d(f_2^k, f_1^{k-1})}{l + u}$$
Since the upper bound of the generalization error of the segmentor $f_1^{k-1}$ is $b_1^{k-1}$, we have $d(f*, \sigma_2) \le \frac{ub_1^{k-1}}{l+u}$. Since $\sigma_2$ contains unlabeled data which may be incorrectly labeled, $\sigma_2$ must be sufficient to guarantee that if the difference of $f_2^k$ and $\sigma_2$ is less than that of $f*$ which means $f_2^k$ "learns" the mistake, then the probability that the generalization error of $f_2^k$ is less than $b_2^k$ is less than $\delta$. Let $M=ub_1^{k-1}$, then the probability that $f_2^k$ has a lower observed difference with $\sigma_2$ than $f*$ is less than
$$P = C_{l+u}^M d(f_2^k, \sigma_2)^M (1-d(f_2^k, \sigma_2))^{l+u-M}$$
Let $b_2^k = \max \{\frac{lb_2^0 + ub_1^0 - u \times d(f_{1-i}^{k-1}, f_{i}^{k})}{l}, 0 \}$,
\begin{align*}
    d(f_2^k, \sigma_2) &= \frac{l\times d(f_2^k, f*) + u\times d(f_2^k, f_1^{k-1})}{l + u}\\
    &\ge \frac{lb_2^k + u\times d(f_2^k, f_1^{k-1})}{l + u}\\
    &\ge \frac{lb_2^0 + ub_1^0}{l + u}
\end{align*}
As the function $C_s^t x^t(1-x)^{s-t}$ is monotonically decreasing in $\frac{t}{s}<x<1$, it follows that
\begin{align*}
P \le C_{l+u}^M (\frac{lb_2^0 + ub_1^0}{l + u})^M (1-\frac{lb_2^0 + ub_1^0}{l + u})^{l+u-M}
\end{align*}
We can approximate the RHS with Poisson Theorem.
\begin{align*}
&C_{l+u}^M (\frac{lb_2^0 + ub_1^0}{l + u})^M (1-\frac{lb_2^0 + ub_1^0}{l + u})^{l+u-M} \\
&\approx \frac{(lb_2^0 + ub_1^0)^M}{M!}e^{-(lb_2^0 + ub_1^0)}
\end{align*}
When $lb_2^0 \le e\sqrt[M]{M!} - M$,
$$
\frac{(lb_2^0 + ub_1^0)^M}{M!}e^{-(lb_2^0 + ub_1^0)} \le e^{lb_2^0}
$$
We show at the beginning that $l \ge \frac{1}{b_2^0}\ln \frac{|\mathcal{H}|}{\delta}$, thus
$$
P \le e^{lb_2^0} \le \frac{\delta}{|\mathcal{H}|}
$$
Given at most $|\mathcal{H}| - 1$ (excluding the optimal $f*$) segmentor with generalization error no less than $b_2^k$ having a lower observed difference with $\sigma_2$ than $f*$ in hypothesis class $\mathcal{H}$, the probability that
$$Pr\big[d(f_2^k, f*) \ge b_2^k \big] \le \delta$$.
In order to let the above derivation holds, we need one more condition which is that the generalization error of $f_1^{k-1}$, which is the counterpart model in the last iteration, is bounded by $b_i^{k-1}$ by probability $\delta$. When $k=0$, which is the initial segmentor that trains on the labeled set only, this condition is satisfied (by supervised PAC learning). When $k=1$, the above holds as the the generalization error of $f_1^0$ is bounded by $b_1^0$ by probability $\delta$. Then, by deduction, we can prove that the above holds for any $k$.
\section{Quantitative Analysis of Homogenization problem}
\label{app:homogenization}
\begin{figure}[H]
\centering
\includegraphics[width=0.4\textwidth]{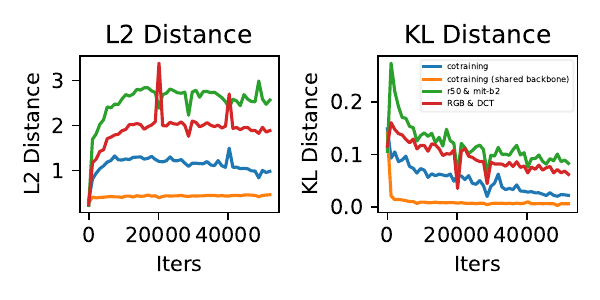} 
\caption{Demonstration of homogenization problem in Co-training} 
\label{fig:kl_l2} 
\end{figure}

To quantitatively analyze the homogenization problem of Co-training (or to quantify the diversity between two models in the Co-training), we further propose two metrics to measure the similarity in target space. As discussed in Section \ref{sec:limitations}, we can only quantify in the target space since measures in parameter space of different architectures is meanless. Specifically, we use L2 distance to measure the similarity of logits output by the two models in Co-training methods.
\begin{equation}
\small
\setlength\abovedisplayskip{0pt}
\setlength\belowdisplayskip{0pt}
D_{l2} = \frac{1}{HWC}\sum_{i=0}^{HW} \sum_{j=0}^{C} \Vert logit^j_{1i}-logit^j_{2i}\Vert_2 \notag
\end{equation}
As the model outputs probabilistic distributions, we can also measure the similarity of models by KL Divergence.
\begin{equation}
\small
\setlength\abovedisplayskip{0pt}
\setlength\belowdisplayskip{0pt}
D_{kl} = \frac{1}{HW}\sum_{i=0}^{HW} \sum_{j=0}^{C} s^j_{1i}\log \frac{s^j_{1i}}{s^j_{2i}} \notag
\end{equation}
As shown in Figure \ref{fig:kl_l2}, we can see that Co-training with a shared backbone suffers the most from the homogenization problem while different architecture and different input domains allow more diverse model in Co-training, which is consistent with the findings in Section \ref{sec:limitations}.
\vspace{-5mm}

\section{Quatification of Diversity in Different Techniques}
After identifying the homogenization problem in Co-training methods, we provide three techniques to alleviate this problem. As discussed in Section \ref{sec:limitations}, \ref{sec:analysis_diversity} and Appendix \ref{app:homogenization}, we first show that the three techniques can reduce the  homogenization (measured by prediction similarity) and then empirically show the effectiveness of each technique individually and combined. Here, we are curious about how much diversity they each introduce, or more specifically, to compare the diversity they bring to the Co-training. We conjecture that with more diversity introduced, the empirical performance is better. The first and simplest approach is to directly and qualitatively analyze the homogenization plots. We can see that different architectures provide more diverse predictions than different input domains as well as Co-training, and Co-training (shared backbone). The second approach can quantify the diversity brought by leveraging one of the three metrics (agree rate, l2, or kld discussed in Appendix \ref{app:homogenization}). Due to the stochastic nature of SGD optimization, we can use an exponential moving average to estimate the metrics. An alternative can be a weighted average of the metrics at the last epoch over the whole dataset. However, we emphasize here that the three techniques tackle homogenization in three different perspectives in the training process and they mutually benefit each other as shown in the ablation study.
\section{VOC PASCAL 2012 Results on ResNet101 and Comparison with SOTA}
\label{app:voc_r101}
We provide comparison with ResNet101 and SegFormer-b3 on \textit{VOC PASCAL 2012} under the second partition protocol mentioned in Section \ref{sec:exp_setup}. For \textit{Diverse Co-training}, we use ResNet101 and SegFormer-b3 as backbones and compare two variants (\ie 2-cps and 3-cps) with other methods with ResNet101 in Table \ref{tab:voc_sota_r101}. We further demonstrate the effectiveness of our \textit{Diverse Co-training} by showing that the improvement over current SOTA methods with resolutions of 321 and 513. We outperform the previous best consistently by more than 2\% with resolution of 321 and around 1\% with resolution of 513. For instance, ours (3-cps) surpasses ST++ \cite{yang2022st++} by 2.8\%, 2.0\% and 2.0\% on 1/16, 1/8 and 1/4 partition protocols respectively. We also compare with AEL \cite{hu2021semi}, U$^2$PL\cite{wang2022semi} and PS-MT \cite{psmt} which obtains the best previous performance. We outperforms the best of them by 0.7\%, 0.8\% and 1.3\% on 1/16, 1/8 and 1/4 partition protocols respectively. It's worth mentioning that, our performance with resolution 321 already outperforms the previous SOTA with resolution 512. The remarkable performance of our \textit{Diverse Co-training} illustrate the significance of diversity in co-training.
\begin{table}[h]
  \small 
  \centering
  \setlength{\tabcolsep}{1mm}{
  \begin{tabular}{c|c|ccc}
    \toprule
    \multirow{2}{*}{Method} & \multirow{2}{*}{Resolution}  & 1/16 & 1/8 & 1/4 \\
    &  & (662) & (1323) & (2646) \\
    \hline
    \hline
    Sup Baseline &  321x321   & 67.5 & 70.4 & 73.7 \\
    CAC \cite{lai2021semi} &  321x321   & 72.4 & 74.6 & 76.3 \\
    CTT*          & 321x321 & 73.7 & 75.1 & -\\
    ST++ \cite{yang2022st++}   & 321x321  & 74.5 & 76.3 & 76.6  \\
    \hline
    ours (2-cps) & 321x321  & \textbf{77.6} & \textbf{78.3} & \textbf{78.7}  \\
    ours (3-cps) & 321x321   & \underline{77.3} & \underline{78.0} & \underline{78.6} \\
    \hline
    \hline
    Sup Baseline & 513x513  & 66.6 & 70.5 & 74.5 \\
    MT \cite{meanteacher}          & 512x512  & 70.6 & 73.2 & 76.6 \\
    CCT \cite{ouali2020semi}   & 512x512  & 67.9 & 73.0 & 76.2\\
    GCT  \cite{ke2020guided}    & 512x512  & 67.2 & 72.2 & 73.6 \\
    CPS \cite{cps}   & 512x512  & 74.5 & 76.4 & 77.7\\
    CutMix \cite{wang2022semi} & 512x512  & 72.6 & 72.7 & 74.3\\
    3-CPS \cite{filipiak2021n}  & 512x512  & 75.8 & 78.0 & 79.0\\
    DSBN‡  & 769x769 & - & 74.1 & 77.8 \\
    ELN  \cite{kwon2022semi} & 512x512  & - & 75.1 & 76.6 \\
    U$^2$PL \cite{wang2022semi}  & 513x513 & 74.4 & 77.6 & 78.7  \\
    PS-MT \cite{psmt} & 512x512  & 75.5 & 78.2 & 78.7  \\
    AEL  \cite{hu2021semi}  & 513x513  & 77.2 & 77.6 & 78.1\\
    \hline
    ours (2-cps) & 513x513 & \textbf{77.9} & \underline{78.7} & \underline{79.0}  \\
    ours (3-cps) & 513x513 & \underline{77.6} & \textbf{79.0} & \textbf{80.0} \\
    
    \bottomrule
  \end{tabular}
  }
  \caption{Comparison with state-of-the-art methods with ResNet101 on the \textit{Pascal VOC 2012} dataset. Labeled images are sampled from the blended training set. Results of MT, CCT, GCT are from \cite{cps}. Results of CTT (denoted by *) is based on DeepLabv2 and results of DSBN (denoted by ‡) is based on Xception65}
  \label{tab:voc_sota_r101}
\end{table}

\section{Detailed DCT Transform}
\label{app:detailed_dct}
We detailed the DCT trasform in this section. As illustrated in Figure \ref{fig:dct_process}, we first transform images to YCbCr color space, consisting of one luma component (Y), representing the brightness, and two chroma components, Cb and Cr, representing the color. Since the spatial resolution of the Cb and Cr channel is reduced by a factor of two, we upsample the original image by two to obtain the same resolution as Y channel. The image is then converted to the frequency domain through DCT transform where each of the three Y, Cb, and Cr channels is split into blocks of 8×8 pixels and transformed to DCT coefficients of 192 channels. 
The two-dimensional DCT coefficients at the same frequency are grouped into one channel to form the three-dimensional DCT cubes. 
After the DCT transform, we obtain frequency domain input of 192 channels but with resolution downsampled by 8. Following \cite{xu2020learning}, we select 64 channels (44, 10 and 10 channels each from Y, Cb and Cr components respectively) close to upper-left squares from the total 192 channels to reduce computation. We refer to \cite{xu2020learning} for more details regarding the channel selections. 

Since the number of channels for frequency domain is different than the RGB domian (\ie three), we have to modify the backbone to adapt it. We take ResNet \cite{resnet} as an example. To be as simple as possible and further reduce training parameters and computation, we remove the stem layers at the beginning of ResNet and modify the first convolution layer in the first ResLayer to have 64 in channels. 

Notice that, the above DCT transform are not contradictory to standard pre-processing techniques widely applied to RGB images it takes RGB images as input, requiring minimum modifications to the current pre-process pipeline and model architecture. To maintain the strong-weak augmentation proposed above, we first perform augmentations on RGB images and then transform it to DCT for training models on the frequency domain. 
\section{Comparison with Knowledge Distillation}
\label{app:kd}
As discussed in Section \ref{sec:ablation}, Co-training is similar to knowledge distillation (KD) in the sense that they both possess a teaching process, the difference lies in that the teacher in KD is usually fixed and teaching is unidirectional while Co-training does not possess the "teacher" and "student" concept and the model teaches each other mutually. To demonstrate that the effectiveness of our method is not simply a knowledge transfer from one model to another but a mutually beneficial process, we compare the knowledge distillation with our method. Specifically, a Segfromer with mit-b2 is trained alone and distills the knowledge to Fixmatch with ResNet50. From Table \ref{tab:kd}, we show knowledge transfer do take effect improving the original FixMatch baseline by 3\%~1\%, which can be attributed to the diverse inductive bias and the high-quality pseudo label introduced by the transformer model. However, we show that our method still outperforms knowledge distillation by ~1\% consistently. This is because Co-training mutually benefits the two models while KD fails to enjoy this benefit. This can be demonstrated from Figure \ref{tab:diff_arch} that Co-training improves the mit-b2 by ~1\% while KD uses a trained and fixed model.
\begin{table}[H]
  \small 
  \centering
  \setlength{\abovecaptionskip}{-0.5mm}
  \setlength{\belowcaptionskip}{-0.5mm}
  \scalebox{1.0}{
  \begin{tabular}{c|c|cccc}
    \toprule
    Method & Param & 1/32 & 1/16 & 1/8 & 1/4 \\
    \hline
    \hline
    FixMatch & 40.5M & 70.28 & 73.36 & 74.0 & 74.3\\
    FixMatch Distill & 65.2M & 74.1 & 74.9 & 75.6 & 75.8\\
    Ours (2-cps) & 65.2M & \textbf{75.2} & \textbf{76.0} & \textbf{76.2} & \textbf{76.5}  \\
    \bottomrule
  \end{tabular}
  }
    \caption{Comparison with knowledge distillation. Labeled images are sampled from the original high-quality training
set.}
  \label{tab:kd}
\end{table}

\section{Detail of Strong Augmentation}
\label{app:augmentation}
We provide a full list of strong augmentations applied in Table \ref{tab:augmentation_list}.
\begin{table*}[h]
  \small 
  \centering
  \begin{tabular}{cl}
    \toprule
    \multicolumn{2}{c}{Weak Augmentation}\\
    \hline
    Random Rescale & Resizes randomly the image by [0.5, 2.0].\\
    Random Flip   & Flip the image horizontally with a probability of 0.5.\\
    Random Crop   & Randomly crop a region from the image.\\
    \hline
    \multicolumn{2}{c}{Strong Augmentation}\\
    \hline
    Color Jitter  & Randomly jitter the color space of the image with a probability of 0.8.\\
    Gaussian Blur & Blurs the image with a Gaussian kernel with a probability of 0.5.\\
    Random Grayscale & Turn the image to greyscale with a probability of 0.2.\\
    Cutmix & Cut a patch from one image and paste the patch to another image. We always apply Cutmix to every image.\\
    \bottomrule
  \end{tabular}
    \caption{List of various image transformations.}
  \label{tab:augmentation_list}
\end{table*}

CutMix is applied twice to the two different views individually. Notably, instead of batch-wise CutMix adopted by CPS \cite{cps, yang2022revisiting}, we use in-batch CutMix which leverages the shuffled samples of the same batch to cutmix. We leverage the random cropped image directly as a weakly augmented view to generate labels. Despite CutMix is applied to each strong view individually, in-batch CutMix allows us to generate cutmixed pseudo labels by forwarding each model only once. 

\section{Number of Parameters}
\label{app:param}
The objective of this section is to (1) demonstrate that our improvement is not trivial by simply adding more parameters and (2) facilitate a fair comparison with the SOTA method. We first report the parameters of the different architectures used in Table \ref{tab:diff_arch}.
\begin{table}[H]
  \small 
  \centering
  \setlength{\tabcolsep}{1mm}{
  \begin{tabular}{c|c}
    \toprule
    Backbone & Param \\
    \hline
    R50 & 2 $\times$ 40.5M = 81M \\
    mit-b2  & 2 $\times$ 24.7M =49.4M \\
    R50 \& mit-b2 & 40.5M + 24.7M = 65.2M \\
    \hline
    ResNeSt50 & 2 $\times$ 42.3M = 84.6M \\
    ResNeXt50 & 2 $\times$ 39.8M = 79.6M \\
    R50 \& ResNeST50 & 40.5M + 42.3M = 82.8M \\
    R50 \& ResNeXT50 &  40.5M + 39.8M = 80.3M \\

    \bottomrule
  \end{tabular}
  }
  \caption{We show the parameters of each architecture.}
  \label{tab:param}
\end{table}
As per Table \ref{tab:param}, our R50 \& mit-b2 possess 20M parameters less than CNN variants such as R50 \& ResNeSt50 and R50 \& ResNeXt50 but still achieve better performance.
Then we compare FixMatch-Distill and FixMatch-Ensemble which uses exactly the same or more parameters than ours but a different learning paradigm. FixMatch-Distill uses a trained Segformer-b2 to distill knowledge to ResNet50 model as described in Appendix \ref{app:kd}. FixMatch-Ensemble is an ensemble of two ResNet50 model is uses 20M parameters more than ours. As shown in the first section of Table \ref{tab:param_com}, our model outperforms both FixMatch-Distill and FixMatch-Ensemble consistently by a large margin. This demonstrates that the improvements by our \textit{Diverse Co-training} is not trivially by adding more parameters. Finally, we also compare the parameters used in our method and the previous SOTA methods. CPS \cite{cps} uses two models to perform Co-training while n-CPS (n=3) \cite{filipiak2021ncps} uses three. Although PS-MT \cite{psmt} uses only one architecture, they leverage two teachers (which are two different sets of parameters) and one student which equals three times the parameters of one model. U$^2$PL  \cite{wang2022semi} leverages the popular teacher-student framework which also leverages two sets of parameters. We show dominant performance with 20M parameters less which further demonstrates the effectiveness of our \textit{Co-training}.
\begin{table}[H]
  \small 
  \centering
  \scalebox{1.0}{
  \begin{tabular}{c|c|cccc}
    \toprule
    Method & Param & 1/32 & 1/16 & 1/8 & 1/4 \\
    \hline
    \hline
    FixMatch Ensemble & 81.0M & 73.0 & 74.3 & 75.6 & 75.9 \\
    FixMatch Distill & 65.2M & 74.1 & 74.9 & 75.6 & 75.8\\
    \hline
    CPS \cite{cps} & 81.0M & - & 72.0 & 73.7 & 74.9 \\
    n-CPS (n=3) \cite{filipiak2021ncps} & 121.5M & - & 72.0 & 74.2 & 75.9 \\
    PS-MT \cite{psmt}   & 121.5M & - & 72.8 & 75.7 & 76.4  \\
    U$^2$PL*  \cite{wang2022semi}  & 81M  & - & 72.0 & 75.1 & 76.2  \\
    \hline
    Ours (2-cps) & 65.2M & \textbf{75.2} & \textbf{76.0} & \textbf{76.2} & \textbf{76.5}  \\
    
    \bottomrule
  \end{tabular}
  }
    \caption{Comparison of parameters and performance with different learning paradigms and previous SOTA. Labeled images are sampled from the original high-quality training
set.}
  \label{tab:param_com}
\end{table}

\section{Visualization}
\begin{figure}[htbp] 
\centering
\includegraphics[width=0.45\textwidth]{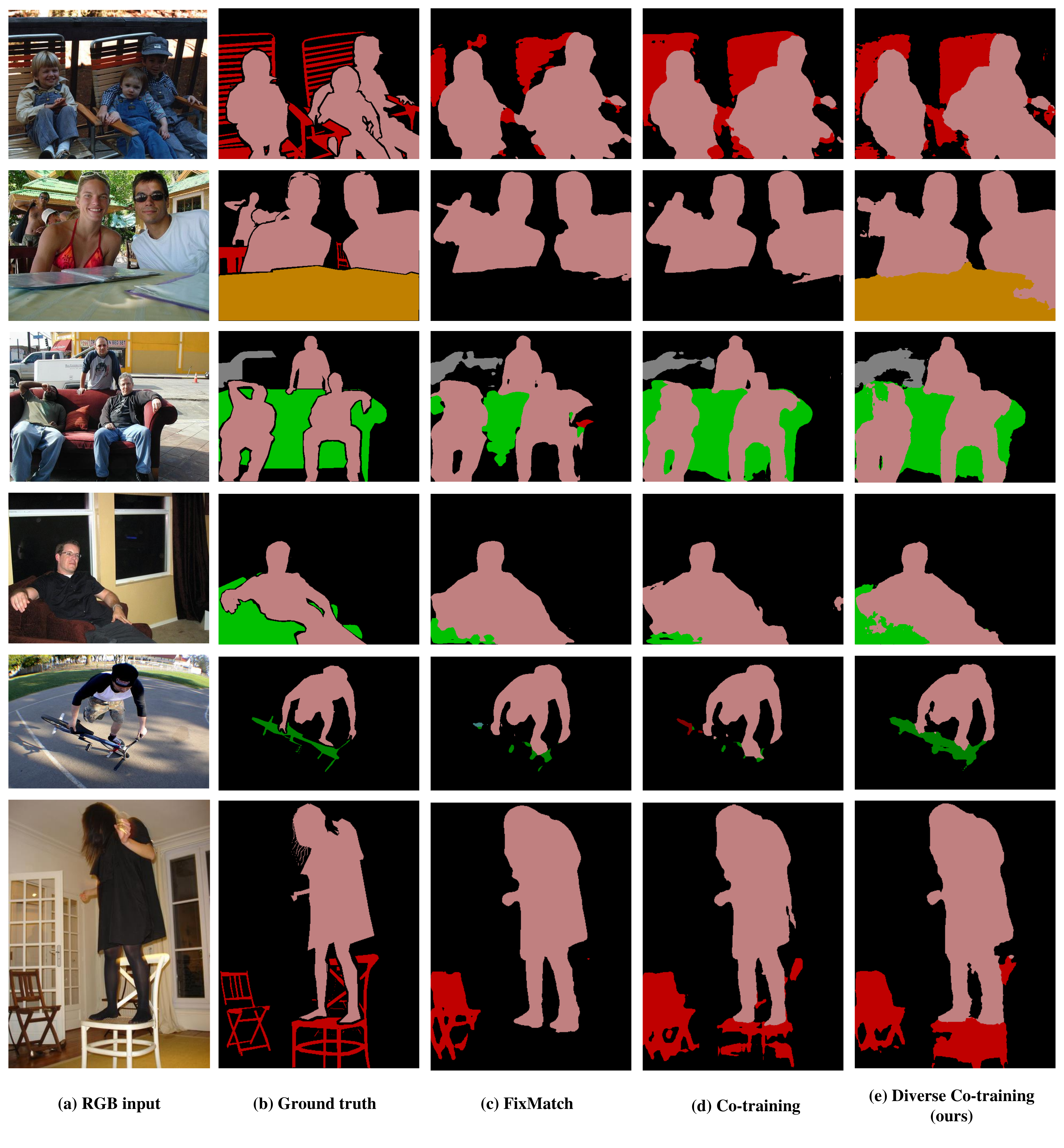}
\caption{Example qualitative results from \textit{PASCAL VOC 2012}. (a) RGB input; (b) ground truth; (c) FixMatch; (d) Co-training baseline; (e) Diverse Co-training (ours). (c) and (d) use DeepLabv3+ with ResNet50 as the segmentation network while (e) uses DeepLabv3+ with ResNet50 and SegFormerb2 (with MLP head) as the two segmentation networks.} 
\label{fig:vis} 
\end{figure}

Figure \ref{fig:vis} visualizes some segmentation results on \textit{PASCAL VOC 2012} validation set. First, we can observe the better results obtained by co-training methods (\ie (d) and (e)) as shown in the third and last row, where FixMatch is prone to under-segmentation (classifies many foreground pixels as background). Our \textit{Diverse Co-training}, compared with co-training baseline, can better segments the small objects that FixMatch and co-training baseline tends to ignore (\eg the forth and fifth row). The FixMatch and co-training baseline tends to ignore some foreground while our \textit{Diverse Co-training} does not, such as the visualization of the second row. These visualization further demonstrate the remarkable performance of \textit{Diverse Co-training} and proves the argument that diversity matters significantly in co-training.

\section{Full Comparison with SOTA on Pascal VOC 2012}
\label{app:full_comparison}
Due to limited space, we only compare the most recent SOTA in Section \ref{sec:comp_sota}. We provide a full comparison here.
\begin{table}[h]
  \small 
  \centering
  \setlength{\tabcolsep}{0.8mm}{
  \begin{tabular}{c|c|ccccc}
    \toprule
    Method & Resolution & 92 & 183 & 366 & 732 & 1464 \\
    \hline
    \hline
    \multicolumn{7}{c}{ResNet50}\\
    \hline
    Sup Baseline & 513x513 & 39.1  & 51.3 & 60.3 & 65.9 & 71.0\\
    PseudoSeg \cite{zou2020pseudoseg} & 512x512 & 54.9 & 61.9 & 64.9 & 70.4 & - \\
    PC$^2$Seg \cite{zhong2021pixel} & 512x512 & 56.9 & 64.6 & 67.6 & 70.9 & - \\
    \hline
    Ours (2-cps) & 513x513  & \underline{71.8} & \underline{74.5} & \textbf{77.6} & \underline{78.6} & \underline{79.8} \\
    Ours (3-cps) & 513x513  & \textbf{73.1} & \textbf{74.7} & \underline{77.1} & \textbf{78.8} & \textbf{80.2}  \\
    \hline
    \hline
    \multicolumn{7}{c}{ResNet101} \\
    \hline
    Sup Baseline & 321x321 & 44.4 & 54.0 & 63.4 & 67.2 & 71.8 \\
    PseudoSeg \cite{zou2020pseudoseg} & 321x321 & 57.6 & 65.5 & 69.1 & 72.4 & 73.2 \\
    PC$^2$Seg \cite{zhong2021pixel} & 321x321 & 57.0 & 66.3 & 69.8 & 73.1 & 74.2 \\
    ReCo \cite{liu2021bootstrapping}     & 321x321  & 64.8 & 72.0 & 73.1 & 74.7 & - \\
    ST++ \cite{yang2022st++}    & 321x321  & 65.2 & 71.0 & 74.6 & 77.3 & 79.1  \\
    \hline
    ours (2-cps) & 321x321  & \underline{74.8} & \textbf{77.6} & \underline{79.5} & \underline{80.3} & \textbf{81.7}  \\
    ours (3-cps) & 321x321  & \textbf{75.4} & \underline{76.8} & \textbf{79.6} & \textbf{80.4} & \underline{81.6}  \\
    \hline
    Sup Baseline & 512x512 & 42.3 & 56.6 & 64.2 & 68.1 & 72.0 \\
    MT \cite{meanteacher}       & 512x512 & 48.7 & 55.8 & 63.0 & 69.16 & - \\
    GCT \cite{ke2020guided}       & 512x512 & 46.0 & 55.0 & 64.7 & 70.7 & - \\
    CTT* \cite{xiao2022semi}      & 512x512 & 64 & 71.1 & 72.4 & 76.1 & - \\
    CPS\cite{cps}      & 512x512  & 64.1 & 67.4 & 71.7 & 75.9 & - \\
    U$^2$PL \cite{wang2022semi}   & 512x512  & 68.0 & 69.2 & 73.7 & 76.2 & 79.5 \\
    PS-MT \cite{psmt}    & 512x512  & 65.8 & 69.6 & 76.6 & 78.4 & 80.0 \\
    \hline
    ours (2-cps) & 513x513  & \textbf{76.2} & \underline{76.6} & \textbf{80.2} & \underline{80.8} & \underline{81.9} \\
    ours (3-cps) & 513x513  & \underline{75.7} & \textbf{77.7} & \underline{80.1} & \textbf{80.9} & \textbf{82.0} \\
    \bottomrule
  \end{tabular}
  }
  \caption{Full Comparison with state-of-the-art methods on the \textit{Pascal} dataset. Labeled images are from the original high-quality training set. Results of CTT (denoted by *) is based on DeeplabV2.}
\end{table}

\begin{table}
  \small 
  \centering
  \setlength{\tabcolsep}{0.8mm}{
  \begin{tabular}{c|c|cccc}
    \toprule
    \multirow{2}{*}{Method} & \multirow{2}{*}{Resolution} & 1/32 & 1/16 & 1/8 & 1/4 \\
    & & (331) & (662) & (1323) & (2646) \\
    \hline
    \hline
    Sup Baseline & 321x321 & 55.8 & 60.3 & 66.8 & 71.3 \\
    CAC\cite{lai2021semi}      &  320x320 & - & 70.1 & 72.4 & 74.0 \\
    ST++\cite{yang2022st++}     & 321x321  & - & 72.6 & 74.4 & 75.4  \\
    \hline
    Ours (2-cps) & 321x321  & \textbf{75.2} & \underline{76.0} & \underline{76.2} & \underline{76.5}  \\
    Ours (3-cps) & 321x321  & \underline{74.9} & \textbf{76.4} & \textbf{76.3} & \textbf{76.6} \\

    \hline
    \hline
    Sup Baseline &  513x513  & 54.1 & 60.7 & 67.7 & 71.9 \\
    CutMix \cite{wang2022semi} & 512x512  & - & 68.9 & 70.7 & 72.5  \\
    CCT \cite{ouali2020semi}    & 512x512  & - & 65.2 & 70.9 & 73.4  \\
    GCT  \cite{ke2020guided}    & 512x512  & - & 64.1 & 70.5 & 73.5  \\
    CPS\cite{cps}     & 512x512  & - & 72.0 & 73.7 & 74.9  \\
    3-CPS \cite{filipiak2021n}  & 512x512  & - & 72.0 & 74.2 & 75.9  \\
    ELN \cite{kwon2022semi}    & 512x512  & - & - & 73.2 & 74.6  \\
    PS-MT \cite{psmt}   & 512x512  & - & 72.8 & 75.7 & 76.4  \\
    U$^2$PL*  \cite{wang2022semi}  & 513x513  & - & 72.0 & 75.1 & 76.2  \\
    \hline
    Ours (2-cps) & 513x513 & \textbf{75.2} & \underline{76.2} & \underline{77.0} & \underline{77.5}  \\
    Ours (3-cps) & 513x513 & \underline{74.7} & \textbf{76.3} & \textbf{77.2} & \textbf{77.7}  \\
    \bottomrule
  \end{tabular}
  }
  \caption{Full Comparison with state-of-the-art methods with ResNet50 on the \textit{Pascal VOC 2012} dataset. Labeled images are sampled from the blended training set. The result of $U^2PL$ is reproduced with the same setting as ours.}
\end{table}
\section{Full Ablation Study}
\label{app:more_ablation}
We further provide a table to show the importance and performance gain of each component. As per table \ref{tab:ablation}, we can see that all component is effective when incorporate into the holistic framework. The combination of diverse domains and different architecture provides the best result of 75.21\%, 75.85\% and 76.23\$ on 1/32, 1/16 and 1/8 labeled data.

\begin{table*}[h]
\centering
\centering
\small
\caption{\small \textbf{Ablation study of different component combinations}
on PASCAL VOC datatset with ResNet50 and SegFormer-b2.
The results are obtained under 1/32, 1/16 and 1/8 partition protocols
and the observations are consistent for
other partition protocols.
$L^s$ represents the supervision loss on the labeled data.
$L^ul$ represents the pseudo supervision loss on the unlabeled data.
SA (strong augmentation) denotes strong-weak augmentation is used. Diff SA stands for different strong augmentation for each model. Diff domain means using RGB and frequency domain to train separate models with cross supervision. Diff arch means different architectures are used to instantiate the two models.
}
\vspace{-2mm}
\label{tab:ablation}
\footnotesize
\begin{tabular}{c|c|c|c|c|c|c|c|c}
\hline 
\multicolumn{6}{c|}{Components}  & \multicolumn{3}{c}{PASCAL VOC} \\ \hline
$\mathcal{L}^s$ & $\mathcal{L}^u$ & SA & diff SA & diff domain & diff arch & 1-32 & 1-16 & 1-8 \\ \hline
\checkmark &  &  &  &  &   & $55.78$ & $60.3$  & $66.79$\\
\checkmark & \checkmark &  &  &  &  & $65.66$ &  $71.28$  & $73.77$\ \\
\checkmark & \checkmark & \checkmark &  &  &  & $70.28$ & $73.36$ & $74.82$  \\
\checkmark & \checkmark &  & \checkmark &  &  & 69.45 & 72.43 &  74.84 \\
\checkmark & \checkmark &  &  & \checkmark &  & 71.58 & 74.94 & 75.97  \\
\checkmark & \checkmark & \checkmark & \checkmark &  &  & $71.07$ & $74.09$ & $74.98$ \\
\checkmark & \checkmark & \checkmark &  & \checkmark & & $72.00$ & $74.10$  & $74.93$  \\
\checkmark & \checkmark & \checkmark &  & & \checkmark & $74.89$ & $75.82$  & $76.08$  \\
\checkmark & \checkmark & \checkmark &  & \checkmark & \checkmark & $\textbf{75.21}$ & $\textbf{75.99}$  & $\textbf{76.23}$  \\ \hline
\end{tabular}
\end{table*}

\end{document}